\documentclass[lettersize,journal]{IEEEtran}
\usepackage{amsmath,amsfonts}
\usepackage{CJKutf8}
\usepackage[linesnumbered,ruled,vlined]{algorithm2e}
\usepackage{graphicx}

\usepackage{array}
\usepackage[caption=false,font=normalsize,labelfont=sf,textfont=sf]{subfig}
\usepackage{textcomp}
\usepackage{stfloats}
\usepackage{url}
\usepackage{verbatim}
\usepackage{graphicx}
\usepackage{cite}
\usepackage{amssymb}
\usepackage{booktabs}
\usepackage{makecell}
\usepackage{multirow}
\usepackage{color}
\usepackage{bbding}
\usepackage{threeparttable}

\begin{document}
\begin{CJK*}{UTF8}{gbsn}


\title{Hallucination Elimination and Semantic Enhancement Framework for Vision-Language Models in Traffic Scenarios}

\author{Jiaqi Fan, Jianhua Wu, Hongqing Chu, Quanbo Ge, Bingzhao Gao
	\thanks{
		
	Jiaqi Fan is with Shanghai Research Institute for Intelligent Autonomous Systems, Tongji University, Shanghai 201804, China (e-mail: fanjq@tongji.edu.cn).
	
	Quanbo Ge is with the School of Automation, Nanjing University of Information Science and Technology, Nanjing 210044, China (e-mail: QuanboGe@163.com).
	
	Jianhua Wu, Hongqing Chu and Bingzhao Gao are with the School of Automotive Studies, Tongji University, Shanghai 201804, China (e-mail: 2332980@tongji.edu.cn; chuhongqing@tongji.edu.cn; gaobz@tongji.edu.cn).
	}
	}

\markboth{Journal of \LaTeX\ Class Files, March~2024}
{Shell \MakeLowercase{\textit{et al.}}: A Sample Article Using IEEEtran.cls for IEEE Journals}

\maketitle

\begin{abstract}
Large vision-language models (LVLMs) have demonstrated remarkable capabilities in multimodal understanding and generation tasks. However, these models occasionally generate hallucinatory texts, resulting in descriptions that seem reasonable but do not correspond to the image. This phenomenon can lead to wrong driving decisions of the autonomous driving system. To address this challenge, this paper proposes HCOENet, a plug-and-play chain-of-thought correction method designed to eliminate object hallucinations and generate enhanced descriptions for critical objects overlooked in the initial response. 
Specifically, HCOENet employs a cross-checking mechanism to filter entities and directly extracts critical objects from the given image, enriching the descriptive text. 
Experimental results on the POPE benchmark demonstrate that HCOENet improves the F1-score of the Mini-InternVL-4B and mPLUG-Owl3 models by 12.58\% and 4.28\%, respectively. Additionally, qualitative results using images collected in open campus scene further highlight the practical applicability of the proposed method.
Compared with the GPT-4o model, HCOENet achieves comparable descriptive performance while significantly reducing costs. 
Finally, two novel semantic understanding datasets, CODA\_desc and nuScenes\_desc, are created for traffic scenarios to support future research. The codes and datasets are publicly available at https://github.com/fjq-tongji/HCOENet.


\end{abstract}


\section{Introduction}

Recently, large vision-language models (LVLMs)~\cite{TITS_1,TITS_2,GPT_4,llava_o1} have achieved remarkable advancements in multimodal tasks, such as image captioning~\cite{TITS_3}, traffic scene understanding~\cite{TITS_4}, end-to-end planning~\cite{TITS_5}, and action reasoning~\cite{Adapt}. 
However, LVLMs occasionally generate responses that deviate from the given image, a phenomenon known as hallucination~\cite{hallucination_survey}. Unlike other fields, traffic scenarios require not only the detection and removal of hallucinations but also the enhancement of descriptions for critical objects overlooked in the initial response. This necessity arises from the complexity of traffic environments and the risk of false negatives, which can lead to erroneous driving decisions and compromise driving safety. Therefore, eliminating hallucinations and enhancing semantic descriptions are novel and challenging problems for LVLMs in traffic scenarios. 


\begin{figure}[!t]
	\centering
	\includegraphics[width=1.0\linewidth]{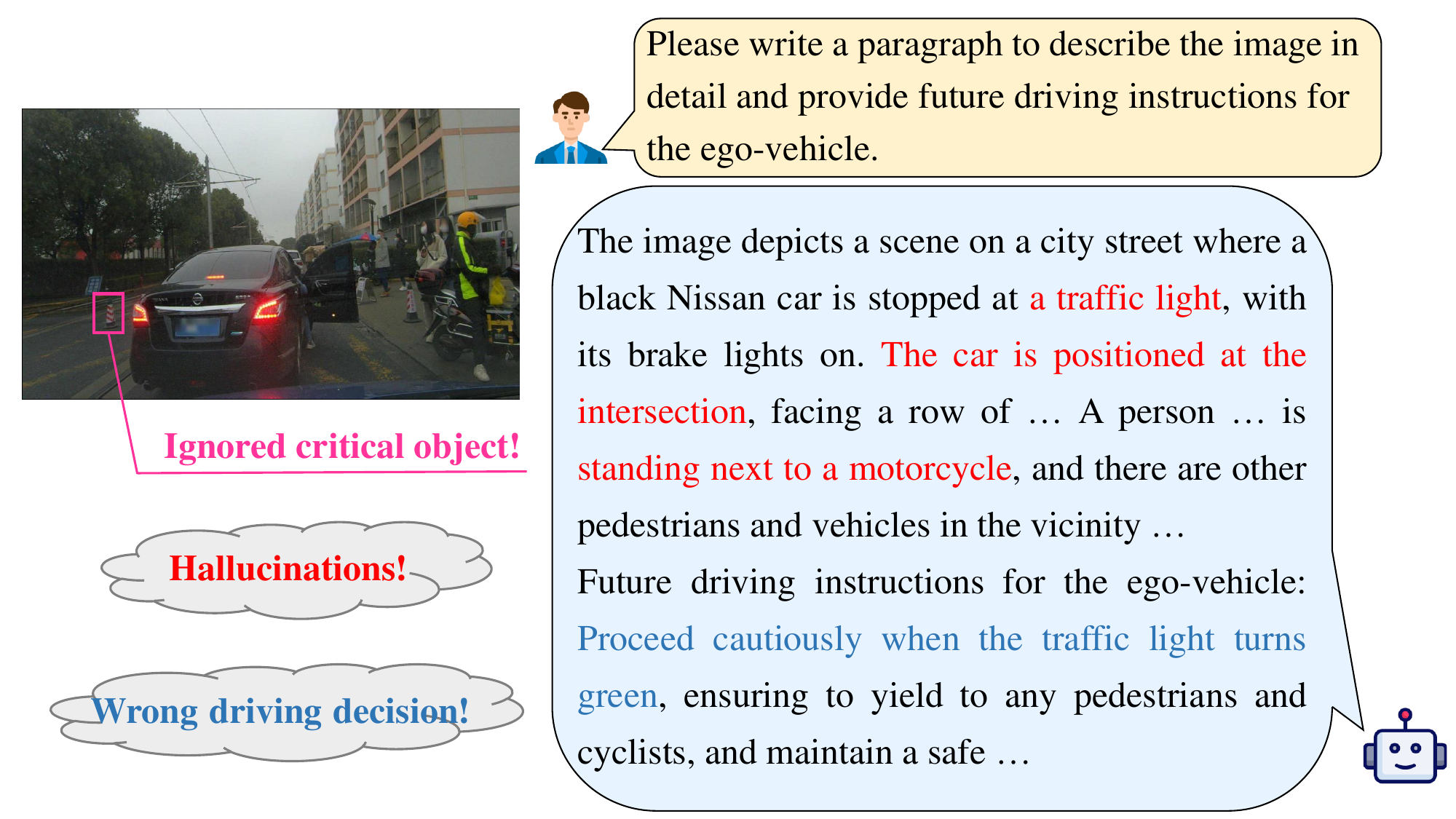}
	\caption{Examples of hallucinations in the traffic scenario, with hallucinatory descriptions highlighted in red, critical object ignored in the description highlighted in pink, and wrong driving decisions highlighted in blue.}
	\label{Fig_introduction_traffic_hallucination}
\end{figure}

The hallucination issue in traffic scenarios can result in wrong driving decisions and disrupt future path planning. Fig.~\ref{Fig_introduction_traffic_hallucination} shows an example of hallucinations in the traffic scenario. In this case, the LVLM mistakenly identifies the scene as an intersection with a traffic light and describes a person standing next to a motorcycle, none of which are present in the image. Correspondingly, the predicted driving instruction erroneously advises the ego-vehicle to wait for the traffic light to turn green before moving forward, which prevents the ego-vehicle from effectively planning its future path and may trap the vehicle in place. Moreover, the response lacks the description of the traffic cone in the left lane, failing to provide an alarm signal for the vehicle to avoid changing lanes to the left, which is dangerous. This example demonstrates the LVLM's limitations in accurately perceiving complex traffic environments, leading to the transmission of misleading information to the autonomous driving system and jeopardizing driving safety. 


The reasons for the occurrence of hallucination phenomenon can be categorized into three aspects. First, LVLMs trained on limited and insufficiently diverse datasets~\cite{LRV_Instruction,hallucidoctor} have constrained the contextual understanding ability, making them prone to misinterpreting the image content and causing hallucinations. Secondly, recent studies~\cite{HACL,RLHF_1} have indicated that a weak vision encoder or an unsatisfactory cross-modal alignment connector can exacerbate the tendency of LVLMs to produce hallucinations. And existing alignment methods such as Q-Former~\cite{BLIP2}, linear projection network~\cite{LLaVA_1.5}, and abstractor module~\cite{mplug_owl2} are insufficient in effectively mapping visual representations into the language space. Thirdly, some works~\cite{VIGC,huang2023survey} attribute the issue to inherent hallucinations in large language models. For instance, the self-attention mechanism overly relies on language priors rather than visual inputs, and the autoregressive generation process causes previously generated tokens to affect subsequent predictions.

Based on the aforementioned reasons for causing hallucinations, many researchers have proposed various strategies to detect and eliminate them. Existing approaches primarily focus on expanding training datasets~\cite{LRV_Instruction,Haloquest_eccv2024}, improving the perception abilities of LVLMs~\cite{IVE,ESREAL}, and integrating expert models for post-hoc calibration~\cite{VDGD,M3ID}. HalluciDoctor~\cite{hallucidoctor} designs a cross-checking paradigm to automatically identify hallucinations and generate counterfactual instruction data, thereby improving the robustness of LVLMs to hallucinations. VCoder~\cite{VCoder} incorporates multiple perception modalities, such as segmentation and depth maps, into LVLMs to improve their perceptual performance. Woodpecker~\cite{woodpecker} employs a multi-step correction framework, leveraging multiple existing multimodal methods to iteratively modify the initial response generated by LVLM. 
However, these methods typically lead to increased model training time and fail to describe objects overlooked in the initial response. 

\begin{figure}[!t]
	\centering
	\includegraphics[width=1.0\linewidth]{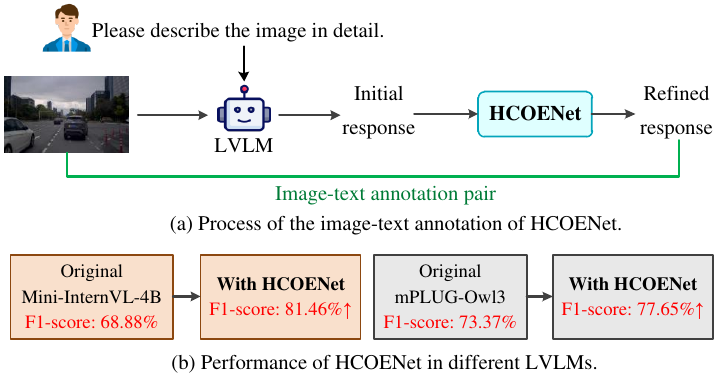}
	\caption{(a) Process of the image-text annotation of HCOENet. (b) The results of HCOENet for eliminating hallucinations on the POPE benchmark. }
	\label{Fig_semantic_labels_CODA_nuscenes}
\end{figure}

For traffic scenarios, in addition to detecting and removing hallucinatory contents, it is equally crucial to perfect descriptions of critical objects ignored in the initial response, such as the traffic cone in Fig.~\ref{Fig_introduction_traffic_hallucination}. To address both challenges, we propose HCOENet, which integrates a hallucination cross-checking framework and a critical-object enhancement framework, delivering an effective solution.
\textcolor{black}{Specifically, the first framework focuses on text descriptions, performing hallucination detection on each entity extracted from individual sentences. For non-existent entities, the corresponding descriptions are then removed.}
\textcolor{black}{The second framework focuses on the given image, directly identifying and describing critical objects within it. Finally, the descriptions from both frameworks are integrated to produce the final refined results.}
This chain-of-thought correction method establishes a clear and logical process for gradually identifying and removing hallucinatory contents while maintaining strong interpretability at each stage.
Additionally, as illustrated in Fig.~\ref{Fig_semantic_labels_CODA_nuscenes}, HCOENet can generate new semantic descriptions for each image after eliminating hallucinations, facilitating the creation of novel scene understanding datasets.

To summarize, the contributions of this paper are summarized as follows:
\begin{itemize}
	\item A chain-of-thought correction method HCOENet is proposed in this paper, which is designed to identify and eliminate semantic hallucinations generated by LVLMs. Meanwhile, the critical-object enhancement framework within HCOENet provides supplementary descriptions for objects overlooked in the initial response, thereby enriching the description and reducing false negatives.
		
	\item Two novel traffic scene semantic understanding datasets, CODA\_desc and nuScenes\_desc, are created using HCOENet. These datasets provide more enrich and hallucination-free semantic descriptions for each driving scenario and are made available to support broader research efforts.


	
	\item The experimental results on ten state-of-the-art LVLMs demonstrate the effectiveness of the proposed HCOENet in eliminating object hallucinations, achieving F1-score improvements of 12.58\%, 4.28\%, and 4.09\% for Mini-InternVL-4B, mPLUG-Owl3, and LLaVA-1.6-13B models, respectively. Furthermore, HCOENet achieves performance comparable to the GPT-4o model at a lower cost. 
	
\end{itemize}

The remainder of this paper is organized as follows. Section \uppercase\expandafter{\romannumeral2} reviews related work of vision-language models and hallucination elimination methods. Section \uppercase\expandafter{\romannumeral3} first introduces the proposed HCOENet method in detail. Then, Section \uppercase\expandafter{\romannumeral4} presents the experimental results across several LVLMs to demonstrate the effectiveness of HCOENet. Finally, Section \uppercase\expandafter{\romannumeral5} creates two semantic scene understanding datasets using HCOENet. Section \uppercase\expandafter{\romannumeral6} is the conclusion of this paper.

\begin{figure*}[!t]
	\centering
	\includegraphics[width=1.0\linewidth]{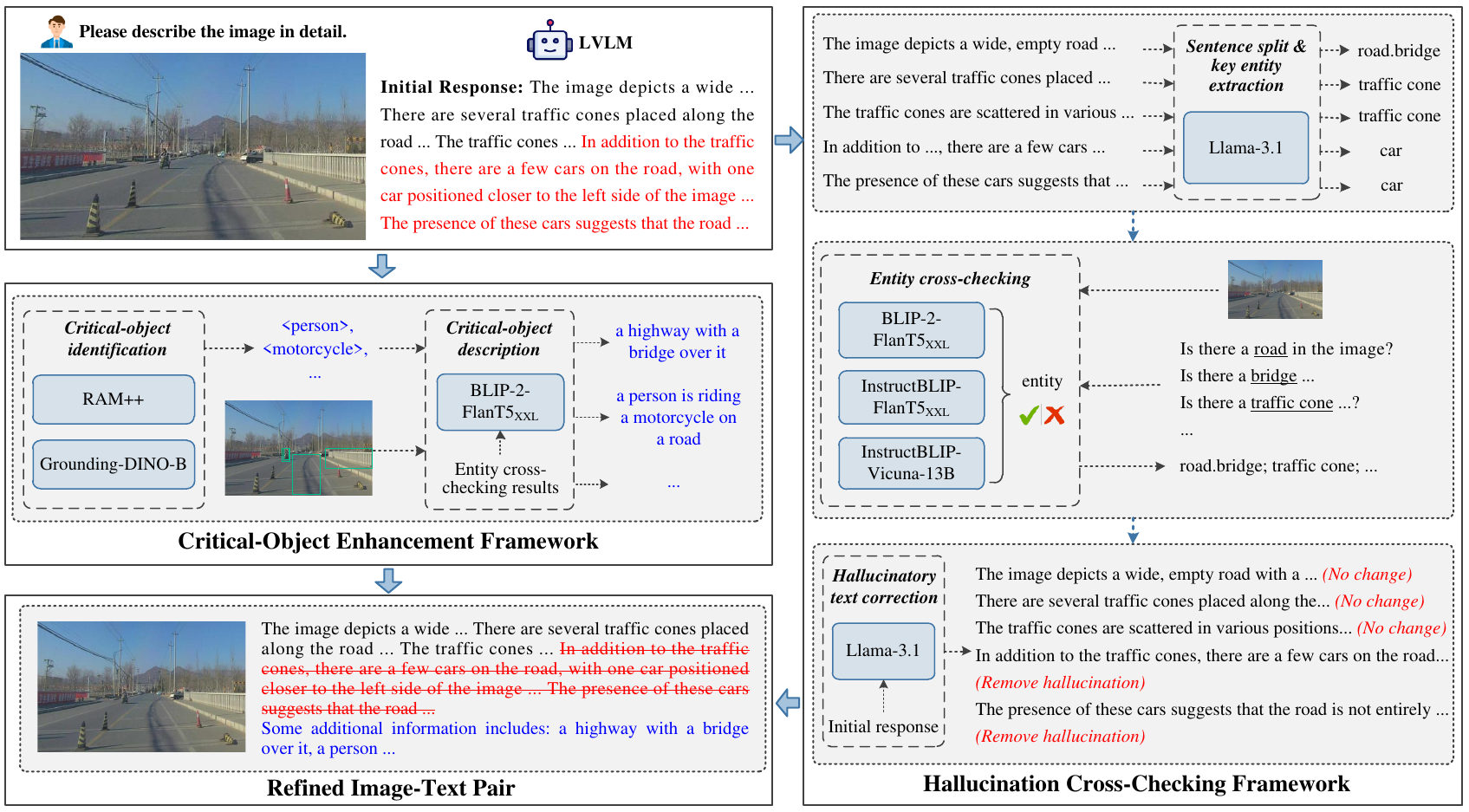}
	\caption{The overall structure of the proposed HCOENet method. Here, the LVLM represents the LLaVA-1.5, mPLUG-Owl2, MiniGPT-4, Mini-InternVL-4B, etc. The hallucinatory texts are highlighted in red, and newly generated semantic contents are highlighted in blue.}
	\label{Fig_overall}
\end{figure*}

\section{Related Work}
\subsection{Large Vision-Language Models}

Recently, an increasing number of LVLMs~\cite{LLaVA_1.6,GPT_4,InternVL_1.5} have showed strong visual comprehension and text reasoning capabilities across many complex tasks, such as image understanding, visual question-answering, and multi-turn conversation. Generally, these LVLMs adopt a structure comprising a visual encoder, a cross-modal alignment network, and a language decoder. Among these components, the cross-modal alignment network plays a crucial role in integrating image features into the language space, which is typically implemented using a linear projector or an adapter module. For the linear projector structure, representative methods include LLaVA-1.5~\cite{LLaVA_1.5}, MiniGPT-4~\cite{MiniGPT4}, mPLUG-Owl3~\cite{mplug_owl3}, InternVL~\cite{InternVL_1.5}, and InternVL2~\cite{InternVL_1.5}. For the adapter structure, BLIP-2~\cite{BLIP2} and InstructBLIP~\cite{InstructBLIP} build a lightweight Q-Former to extract the most relevant visual information for the language generator. Meanwhile, mPLUG-Owl2~\cite{mplug_owl2} introduces a modality-adaptive approach to project visual and linguistic features into a shared semantic space, and Qwen-VL~\cite{Qwen_vl} proposes a randomly initialized cross-attention module as the adapter network.


However, as the length of their output response increases, these LVLMs are prone to generating semantic hallucinations, producing descriptions that appear plausible but are factually incorrect. These defects lead to weak cross-modal understanding ability of LVLMs for the current scene, highlighting the urgent need to detect and eliminate hallucinations.

\subsection{Hallucination Elimination Methods}
To address the hallucinatory issues caused by LVLMs, recent studies have proposed various solutions, focusing on data expansion, model performance, and inference optimization. 

First, some researchers mitigate hallucinations by enriching the training datasets. The LRV-Instruction~\cite{LRV_Instruction} dataset contains both positive and negative instruction samples across different semantic levels. By fine-tuning the model on such dataset, the hallucinations of MiniGPT-4 and mPLUG-Owl are greatly reduced. Recaption~\cite{Recaption} employs the ChatGPT model to rewrite the captions in the dataset, followed by fine-tuning LVLMs on the updated dataset, effectively minimizing fine-grained hallucinations. VIGC~\cite{VIGC} adopts a visual instruction generation approach to create more visual question-answer pairs. This method further integrates an iterative Q-Former update strategy during inference, which helps refine and correct inaccurate descriptions, enhancing the reliability of outputs.




Then, some researchers reduce hallucinations by enhancing the perceptual ability of the visual encoder or the generative ability of the language model. IVE~\cite{IVE} incorporates multiple visual expert models, such as object detectors and OCR, into the LVLM to enrich the model's stored knowledge. LION~\cite{LION} devises a novel visual knowledge-enhanced LVLM, which injects spatial-aware knowledge and high-level semantic visual evidence into the large language model, thereby enhancing the multimodal understanding capability. HACL~\cite{HACL} introduces contrastive learning to the LVLM and utilizes hallucinatory contents as hard negative samples to better align visual and textual representations, reducing misalignment errors. These two types of methods heavily depend on the quality of the constructed instruction data or require significant hardware resources. Compared with them, our method is a training-free architecture that eliminates the need to collect extensive training data, making it easier to aggregate into more LVLMs.



In addition, some works~\cite{IBD,HALC,LURE} propose post-processing methods in the inference or decoding stage to eliminate hallucinations. VCD~\cite{VCD} performs a novel visual contrastive strategy in the decoding process, which compares the output distributions from original and distorted visual inputs. This method ensures stronger consistency between the generated contents and given image. LogicCheckGPT~\cite{LogicCheckGPT} devises a logical closed loop method, which is implemented by constructing object-to-attribute inquiring and attribute-to-object inquiring to ensure the logical consistency. Volcano~\cite{Volcano} builds a feedback-guided revision model, following a program of iterative critique-revision-decide steps.

\textcolor{black}{These methods typically focus on detecting and removing direct hallucinations in the descriptive texts generated by LVLMs but fail to provide supplementary descriptions for key objects overlooked in the initial response. In contrast, our method not only eliminates hallucinatory contents but also directly extracts objects from the image to enrich semantic descriptions, resulting in better alignment between image and text modalities. }
		


\section{Method}
\subsection{Overall Framework}
The overall architecture of the proposed HCOENet is illustrated in Fig.~\ref{Fig_overall}, which consists of two main components: a hallucination cross-checking framework and a critical-object enhancement framework. The hallucination cross-checking framework focuses on analyzing each sentence in the initial response generated by the LVLM to identify and remove hallucinatory contents. And the critical-object enhancement framework focuses on directly extract critical traffic elements from the given image, which is specifically tailored for traffic scenarios to reduce false negatives and improve the completeness of descriptive results.
	
Specifically, when equipped solely with the hallucination cross-checking framework, the resulting HCNet employs three off-the-shelf vision-language models to filter entities. Correspondingly, the initial response generated by LVLMs are corrected based on the filtered results, ensuring greater alignment with the actual visual contents. The chain-of-thought correction method in the HCNet can effectively identify and eliminate object hallucinations. However, some critical elements in the image, such as small-sized motorcycles or persons, may be overlooked in the initial response, leading to an excess of false negatives. To address this, a critical-object enhancement framework is introduced for supplementary information. Together, HCNet and critical-object enhancement framework form the unified HEORNet model.


\subsection{Sentence Split and Key Entity Extraction}

Given that the initial response generated by the vision-language model is relatively lengthy and contains multiple sub-sentences, each potentially including hallucinatory outputs, our method adopts a systematic approach. First, the long descriptions are divided into several short sentences, followed by sequential corrections for hallucinations in each clause. Explicitly, we employ a pre-trained spaCy model to break up the initial captions into individual sentence fragments. Then, a large language model Llama-3.1-8B~\cite{llama3} is utilized to extract entity words from each sentence, which refer to the objects of specific categories, excluding adjectives and verbs. Fig.~\ref{Fig_entity_extraction}(a) shows the designed few-shot prompts, in which we fully consider that fundamentally similar objects should be merged into the same category, and each entity word should be output in the singular form. For instance, the sentence ``\textit{There are two cars in the foreground, one of which is a blue car, and another car is located further back on the street.}" contains two entity words \textit{car} and \textit{street}.
	
In addition, the prompts specify that abstract or non-specific entity words should not be extracted, such as the word ``environment" in the sentence ``\textit{The overall atmosphere of the image suggests a busy urban environment}" is not extracted, and ``None" is returned in this case. Moreover, the few-shot prompts include some examples showing the expected outputs for various input sentences. These examples serve to instruct the large language model effectively, ensuring the generated outputs align with the intended requirements.

\subsection{Entity Cross-Checking and Hallucination Correction}


\begin{algorithm}[!t]
	\caption{Entity Cross-Checking}
	\label{Algorithm_1_entity_filtering_three_blip_models}
	\LinesNumbered
	\KwIn{A given image $img$, list of entity words extracted from the captions of the given image $entity\_lst$     \\
	}
	\KwOut{Entity cross-checking results $filter\_result$}
	\For{$w_i$ in $entity\_lst$}{
		$pro$ $\leftarrow$ Is there a $w_i$ in the image? Please answer only with yes or no;  \\
		$ans_1$ $\leftarrow$ BLIP-2-$\text{FlanT5}_{\mathrm{XXL}}$($pro$, $img$); \\
		$ans_2$ $\leftarrow$ InstructBLIP-$\text{FlanT5}_{\mathrm{XXL}}$($pro$, $img$); \\
		\uIf{$no$ in $ans_1$ and $no$ in $ans_2$}{
			$filter\_result$.append(None);
		}
		\uElseIf{$yes$ in $ans_1$ and $yes$ in $ans_2$}{
			$filter\_result$.append($w_i$);
		}
		\Else{
			$ans_3$ $\leftarrow$ InstructBLIP-Vicuna-13B($pro$, $img$); \\
			\eIf{$yes$ in $ans_3$}{
				$filter\_result$.append($w_i$);
			}{$filter\_result$.append(None)}    
		}
	}
\end{algorithm}

HCOENet employs a cross-checking method to verify whether each extracted entity word exists in the image, detailed in Algorithm~\ref{Algorithm_1_entity_filtering_three_blip_models}. First, BLIP-2-$\text{FlanT5}_{\mathrm{XXL}}$~\cite{BLIP2} is utilized, and the prompt is ``\textit{Is there a \{word\} in the image? Please answer only with yes or no.}" Then, InstructBLIP-$\text{FlanT5}_{\mathrm{XXL}}$~\cite{InstructBLIP} is also utilized to check the entity word, and the prompt is the same. If both models respond with \textit{no}, it concludes that the entity does not exist, returning ``None". Conversely. if both models answer with \textit{yes}, the entity is considered to exist and should be reserved. Explicitly, for cases where the models produce conflicting responses, for example, the answer of BLIP-2-$\text{FlanT5}_{\mathrm{XXL}}$ model is \textit{yes} and the answer of InstructBLIP-$\text{FlanT5}_{\mathrm{XXL}}$ model is \textit{no}, a larger model InstructBLIP-Vicuna-13B~\cite{InstructBLIP} is consulted to make the final judgment. If this larger model responds with \textit{yes}, the entity is deemed present and should be retained. Otherwise, ``None" is returned. This method of using multiple models for entity verification effectively mitigates errors caused by hallucinations in any multimodal model and maximizes the accuracy of entity cross-checking process.

\begin{figure}[!t]
	\centering
	\includegraphics[width=1.0\linewidth]{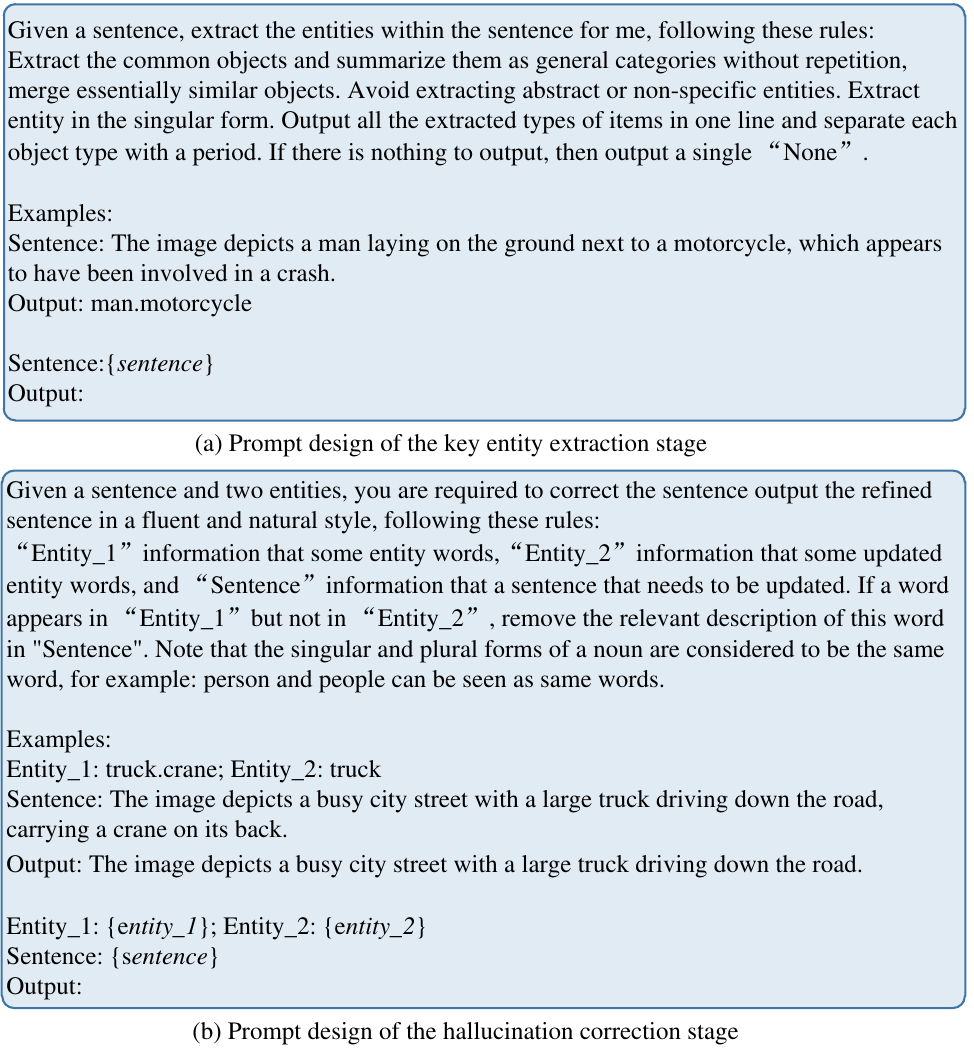}
	\caption{Few-shot prompting design for the key entity extraction stage and hallucination correction stage.  } 
	\label{Fig_entity_extraction}
\end{figure}

Based on the entity update results through the above process, the corresponding sub-sentence needs to be refined and rewritten. In detail, each sentence is corrected using the Llama-3.1-8B~\cite{llama3} model,  guided by the few-shot prompts illustrated in Fig.~\ref{Fig_entity_extraction}(b). These prompts are designed to instruct the language model to obtain refined descriptions free from hallucinatory contents by incorporating specific correction rules and examples. In the refinement process, three inputs are provided to the large language model: the initially extracted entity words ``entity\_1", the filtered entity words ``entity\_2", and the descriptive sentence generated by the LVLMs. For words present in ``entity\_1" but absent in ``entity\_2", they are identified as hallucinatory and have already been filtered out during the entity cross-checking stage. Conversely, the relevant descriptions in the input sentence should also be removed. More precisely, if ``entity\_1" and ``entity\_2" are identical, it indicates that the sentence contains no hallucinations and does not require modification. 

\subsection{Critical-Object Enhancement Framework}

\begin{algorithm}[!t]
	\caption{Critical-Object Identification and Description}
	\label{Algorithm_2_object_detection_describe}
	\LinesNumbered
	\KwIn{A given image $img$, bounding box threshold $\alpha$, entity cross-checking results $filter\_result$
	}
	\KwOut{RAM++ tagging results $obj\_lst_1$, Grounding-DINO-B detection results $obj\_lst_2$, descriptions of detected objects $obj\_lst\_desc$}    
	$obj\_lst_1$ $\leftarrow$ RAM++($img$);    \\
	\For{$w_j$ in $obj\_lst_1$}{
		$output$ $\leftarrow$ Grounding-DINO-B($img$, $w_j$);    \\
		\eIf{$output.logit$ $>=$ $\alpha$}{
			$obj\_lst_2$.append($w_j$);  \\
			\eIf{$w_j$ not in $filter\_result$}{
				$pro$ $\leftarrow$ Describe the $w_j$ in the image with only one sentence;  \\
				$cap_j$ $\leftarrow$ BLIP-2-$\text{FlanT5}_{\mathrm{XXL}}$($pro$, $img$);   \\
				$obj\_lst\_desc$.append($cap_j$);
			}{\textbf{break}}
			
		}{\textbf{break}}
	}
\end{algorithm}

The response generated by LVLMs generally only focuses on the most prominent objects in the image, while neglecting smaller or more distant objects, such as pedestrians, motorcycles, and the traffic cone in Fig.~\ref{Fig_introduction_traffic_hallucination}. However, overlooking these objects can lead to false negatives, which pose significant risks to driving decisions, especially in traffic scenarios. 

To address this issue, we introduce the open-set tagging method RAM++~\cite{RAM++} and open-set detection method Grounding-DINO-B~\cite{GroundingDINO} in the critical-object enhancement framework to directly extract traffic-related objects directly from the image. The detailed critical-object identification and description process is illustrated in Algorithm~\ref{Algorithm_2_object_detection_describe}. First, the RAM++ model is utilized to identify as many objects as possible in the image, focusing on the elements related to traffic scenarios such as pedestrians, motorcycles, vehicles, and trucks. For each recognized tag, the Grounding-DINO-B model, known for its higher precision, is employed to locate the object and verify its existence. Here, the bounding box threshold $\alpha$ is set to 0.35, meaning that only objects with the detection score exceeding $\alpha$ are retained. By combining the results of tagging and detection models, this stage ensures accurate and reliable results of relevant objects in traffic scenarios.

Here, only objects that are not identified in the entity cross-checking stage need to be detected and described. More precisely, the BLIP-2-$\text{FlanT5}_{\mathrm{XXL}}$~\cite{BLIP2} method is employed to describe these newly detected objects, with the prompt of ``\textit{Describe the \{object\} in the image with only one sentence.}" Otherwise, if the detected object has already appeared in the entity cross-checking results, it will not be described again. Finally, the newly generated descriptions, prefixed with ``\textit{Some additional information includes:}", are combined with the corrected response to form the final results.

\section{Experiments}
Section~\uppercase\expandafter{\romannumeral4}-A introduces the dataset, vision-language models, and hallucination evaluation benchmark used in our work. Section~\uppercase\expandafter{\romannumeral4}-B shows ablation studies conducted on the CODA dataset. Section~\uppercase\expandafter{\romannumeral4}-C presents hallucination evaluation results of several LVLMs on the POPE benchmark. Finally, Section~\uppercase\expandafter{\romannumeral4}-D discusses the comparison results, inference time, and qualitative results. 

\subsection{Experimental Details and Benchmark}  
\subsubsection{Dataset}  
In this paper, we evaluate the performance of the proposed HCOENet using two traffic datasets CODA and nuScenes. First, we conduct extensive ablation studies and comparison experiments on the CODA~\cite{CODA} dataset to assess the model's ability for alleviating hallucinations. The CODA dataset focuses on corner case scenarios, containing 4884 images in both validation and test sets, collected from 1500 carefully selected real-world driving scenes. The corner case objects in the dataset can be divided into seven classes: vehicle, pedestrian, cyclist, animal, traffic facility, obstruction, and others. Then, the HCOENet is extended to an automatic annotation pipeline, generating annotations for images in the CODA and nuScenes datasets. Here, the nuScenes dataset consists of 1000 scenes, each with a duration of 20 seconds, and contains more than 40000 images collected from Boston and Singapore. 


\subsubsection{LVLMs}  
All the LVLMs used in this paper are summarized in Table~\ref{Table_introduce_vlms}. For the baseline methods, we select five mainstream models with varying sizes, including MiniGPT-4~\cite{MiniGPT4} using Vicuna-v0 with 7.0 billion parameters as the language model, LLaVA-1.5-7B~\cite{LLaVA_1.5} using Vicuna-1.5 with 7.0 billion parameters as the language model, LLaVA-1.5-13B~\cite{LLaVA_1.5} using Vicuna-v1.5 with 13 billion parameters as the language model, mPLUG-Owl2~\cite{mplug_owl2} using LLaMA with 7.0 billion parameters as the language model, and Mini-InternVL-4B~\cite{InternVL_1.5} using InternViT as the image encoder and Phi-3-mini as the language decoder with approximately 4.0 billion parameters. For the comparison methods, eight additional models are employed, including LLaVA-1.6-7B~\cite{LLaVA_1.6} using Vicuna with 7.0 billion parameters as the language model, LLaVA-1.6-13B~\cite{LLaVA_1.6} using Vicuna with 13 billion parameters as the language model, mPLUG-Owl~\cite{mplug_owl} using LLaMA as the  language model, mPLUG-Owl2.1~\cite{mplug_owl2} using Qwen with 7.0 billion parameters as the language model, mPLUG-Owl3~\cite{mplug_owl3} using Qwen2 as the language model, InternVL2-4B~\cite{InternVL_1.5} using Phi‑3‑mini model as the language model, InternVL2-8B~\cite{InternVL_1.5} using InternLM2 with 7 billion parameters as the language model, and InternVL2-40B~\cite{InternVL_1.5} using Nous-Hermes-2-Yi with 34 billion parameters as the language model. For each multimodal model, the traffic scene image, along with the prompt ``\textit{Please describe the images in detail}", is used as input to generate descriptions of the current scenario. In this work, all the experiments are conducted using four NVIDIA A800 GPUs. 


\begin{table}[!t]
	\renewcommand{\arraystretch}{1.1}
	\caption{Summary of all LVLMs used in this paper. $\theta$ represents model parameters, and ``B" represents billion.}
	\centering
	\begin{tabular}{@{}m{2.65cm}|m{0.8cm}|m{0.53cm}|m{1.57cm}m{1.47cm}}
		\hline
		LVLM  &Release Time   &$\theta$   &Visual Encoder   &Language Decoder     \\
		\hline
		\cmidrule{1-5}
		mPLUG-Owl~\cite{mplug_owl}  &2023.05    &7.2B  	&ViT-L     &LLaMA     \\
		MiniGPT-4~\cite{MiniGPT4}  &2023.08    &8.0B  	&ViT-G    &Vicuna-7B     \\
		LLaVA-1.5-7B~\cite{LLaVA_1.5}  &2023.10    &7.3B  	&CLIP ViT-L    &Vicuna-7B     \\
		LLaVA-1.5-13B~\cite{LLaVA_1.5}  &2023.10    &13.3B  	&CLIP ViT-L    &Vicuna-13B     \\
		mPLUG-Owl2~\cite{mplug_owl2}  &2023.11    &8.2B  	&ViT-L     &LLaMA     \\
		LLaVA-1.6-7B~\cite{LLaVA_1.6}  &2024.01     &7B      &CLIP-ViT    &Vicuna-7B   \\ 
		LLaVA-1.6-13B~\cite{LLaVA_1.6}  &2024.01     &13B      &CLIP-ViT    &Vicuna-13B   \\ 
		mPLUG-Owl2.1~\cite{mplug_owl2}  &2024.02     &9.8B    &ViT-G   &Qwen    \\	
		Mini-InternVL-4B~\cite{InternVL_1.5}  &2024.05    &4.2B  	&InternViT    &Phi-3-mini     \\
		InternVL2-4B~\cite{InternVL_1.5}  &2024.07    &4.2B   	&InternViT     &Phi‑3‑mini      \\
		InternVL2-8B~\cite{InternVL_1.5}  &2024.07    &8B  	&InternViT    &InternLM2     \\
		InternVL2-40B~\cite{InternVL_1.5}  &2024.07    &40B  	&InternViT    &Nous-Hermes-2-Yi     \\
		mPLUG-Owl3~\cite{mplug_owl3}  &2024.08     &8.0B    &SigLIP-400M   &Qwen2    \\           
		\cmidrule{1-5}
	\end{tabular}

	\label{Table_introduce_vlms}
\end{table}

\subsubsection{Hallucination Evaluation Benchmark} 

\begin{figure}[!t]
	\centering
	\includegraphics[width=1.0\linewidth]{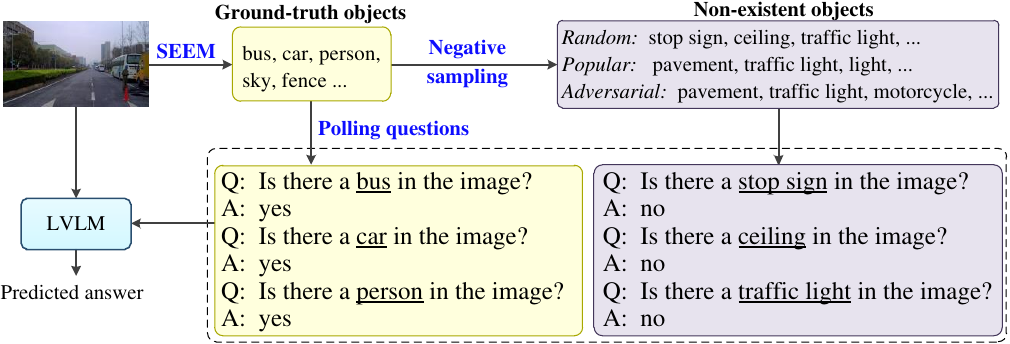}
	\caption{Hallucination evaluation process of the POPE benchmark.}
	\label{Fig_POPE}
\end{figure}

The polling-based object probing evaluation (POPE)~\cite{POPE} is a widely used benchmark for evaluating object hallucination. In this benchmark, the evaluation of object hallucinations is considered as a binary classification task, prompting the LVLMs to generate either \textit{yes} or \textit{no} for a designed question. The specific evaluation process is shown in Fig.~\ref{Fig_POPE}. For each input image, six question-answer pairs are created, with three answers being \textit{yes} and three being \textit{no}, \textit{e.g.,} ``\textit{Is there a car in the image? yes.}" Here, questions with answer \textit{yes} are constructed based on ground-truths or objects extracted from the automatic segmentation method SEEM~\cite{SEEM}, while the questions with answer \textit{no} are constructed by sampling from negative objects. In particular, three different sampling strategies are utilized including random sampling, popular sampling, and adversarial sampling. Then, the questions formulated above, along with the given image, are fed into the LVLMs to obtain predicted answers including \textit{yes} or \textit{no}. And the generated answers are then compared with the ground-truth answers to assess the degree of hallucinations. In  this paper, all the ablation studies are conducted using the LLaVA-1.5-7B model under the popular sampling.	 

Unlike the POPE benchmark which only uses 500 images from the MSCOCO dataset as inputs, all images from the CODA dataset are employed as inputs in this work. In the designed questions, only words related to traffic scenarios are included, while unrelated terms, such as tree, sky, and fence are filtered out. Besides, the descriptions before and after correction, along with the designed questions are provided as prompts to the LVLM. During the process of hallucination assessment, multiple metrics are employed, including precision, recall, F1-score, accuracy, and Yes rate.

\begin{table*}[!t]
	\renewcommand{\arraystretch}{1.0}
	\caption{Ablation studies of the effectiveness of each stage in the HCOENet. Stage1 refers the sentence split and key entity extraction, stage2 refers the entity cross-checking, stage3 refers the hallucination correction, stage4 refers the critical-object identification, stage5 refers the object description, and stage6 refers integrating descriptions from two frameworks. (\%)}
	\centering
	\begin{tabular}{@{}m{2.2cm}|m{0.55cm}<{\centering}m{0.6cm}<{\centering}m{0.6cm}<{\centering}m{1.0cm}<{\centering}m{2.5cm}<{\centering}m{0.5cm}<{\centering}m{0.7cm}<{\centering}|m{0.95cm}<{\centering}m{0.6cm}<{\centering}m{0.6cm}<{\centering}m{1.05cm}<{\centering}|m{0.9cm}<{\centering}}
		\hline
		Method  &Stage1   &Stage2   &Stage3  &Stage4 (RAM++)  &Stage4 (Grounding-DINO-B) &Stage5  &Stage6  &Precision &Recall  &F1-score &Accuracy   &Average  \\
		\hline
		\cmidrule{1-13}
		Baeline &  &    &    &    &     &     &    &59.77    &84.08    &69.87    &63.75   &69.37       \\
		Model A (HCNet)  &\Checkmark &\Checkmark  &\Checkmark  &  &   &   &  &\textbf{64.35} 	&80.53 	&71.53 	&67.95  &71.09      \\   
		Model B  & &  &  &\Checkmark   &   &\Checkmark   &\Checkmark  &52.60 	&\textbf{94.33} 	&67.55 	&54.67  &67.29      \\   
		Model C &   &  &  &\Checkmark   &\Checkmark   &\Checkmark   &\Checkmark  &59.66 	&90.13 	&71.80 	&64.60   &71.55        \\   
		Model D  &\Checkmark &\Checkmark  &\Checkmark  &\Checkmark   &   &\Checkmark   &\Checkmark  &52.69 	&94.15 	&67.57 	&54.80   &67.30      \\
		Ours (HCOENet)  &\Checkmark &\Checkmark  &\Checkmark  &\Checkmark   &\Checkmark   &\Checkmark   &\Checkmark  &62.79 	&88.30 	&\textbf{73.39} 	&\textbf{67.98}    &\textbf{73.11}    \\
		\cmidrule{1-13}
	\end{tabular}
	
	\label{Table_Ablation_1}
	
\end{table*}

\begin{table*}[!t]
	\renewcommand{\arraystretch}{1.0}
	\caption{Ablation studies of different model in the entity cross-checking process. (\%)}
	\centering
	\begin{tabular}{@{}m{2.2cm}|m{1.9cm}<{\centering}m{1.9cm}<{\centering}m{1.9cm}<{\centering}|m{1.2cm}<{\centering}m{1.3cm}<{\centering}m{1.3cm}<{\centering}m{1.3cm}<{\centering}|m{1.3cm}<{\centering}}
		\hline
		Method   &BLIP-2 $\text{FlanT5}_{\mathrm{XXL}}$     &InstructBLIP $\text{FlanT5}_{\mathrm{XXL}}$      &InstructBLIP Vicuna-13B     &Precision  &Recall  &F1-score    &Accuracy   &Average \\
		\hline
		\cmidrule{1-9}
		Model C &  &    &    &59.66   &\textbf{90.13} 	&71.80 	&64.60   &71.55     \\
		Model E &\Checkmark  &    &     &62.95 	&87.50 	&73.22 	&68.00   &72.92     \\
		Model F &\Checkmark  &\Checkmark    &     &\textbf{63.16} 	&87.52 	&73.37 	&\textbf{68.24}   &73.07    \\
		Model G   &  &    &\Checkmark     &61.68 	&85.64 	&71.63 	&66.22     &71.29     \\
		Model H &\Checkmark  &    &\Checkmark     &61.03 	&89.04 	&72.42 	&66.09    &72.14   \\
		Model I &  &\Checkmark    &\Checkmark     &61.01 	&89.07 	&72.42 	&66.07   &72.14    \\
		Ours (HCOENet) &\Checkmark  &\Checkmark    &\Checkmark     &62.79 	&88.30 	&\textbf{73.39} 	&67.98  &\textbf{73.11}     \\
		\cmidrule{1-9}
	\end{tabular}
	\label{Table_Ablation_2}
\end{table*}

\subsection{Ablation Studies}
\subsubsection{Different Stages}
In this subsection, we evaluate the effectiveness of each stage in the proposed HCOENet, which is reported in Table~\ref{Table_Ablation_1}. Here, the Model A, Model B, Model C, and Model D are our method with different correction stages. The baseline method involves directly questioning the initial response generated by the LVLMs. Since these descriptions often contain numerous hallucinatory elements, the evaluation metrics are notably low. In model A, the HCNet model substantially improves the precision score, achieving the highest of 64.35\%.  
	
Model B and model C do not attempt to detect or remove hallucinations in the initial response. Instead, they directly extract entity information from the given image using two open-set models. These methods can greatly reduce false negatives and improve recall values, reaching the highest 94.33\%. However, due to the unresolved hallucinatory errors, the overall performance is still not optimal. Meanwhile, these experiments demonstrate the effectiveness of Grounding-DINO-B module, which refines the detection results from the RAM++ module, thereby enhancing the detection accuracy to a certain extent, \textit{e.g.} increasing the model accuracy by 9.93\% and F1-score by 4.25\%. In model D, only the RAM++ method is utilized to detect objects in the image, without incorporating corrections from Grounding-DINO-B. This results in the introduction of substantial erroneous information and numerous false-positive samples. Consequently, the model suffers from a pretty high recall value but the lowest precision value. Finally, when all components are integrated into the proposed framework, the HCOENet effectively detects and eliminates hallucinations in the original descriptions while enriching the semantic contents. As a result, the model achieves the highest F1-score of 73.39\% and model accuracy of 67.98\%.

\subsubsection{Different Cross-Checking Methods}



In this subsection, we evaluate the effectiveness of different methods for entity cross-checking, with the results reported in Table~\ref{Table_Ablation_2}. Here, the Model E, Model F, Model G, Model H, and Model I are our method with different correction stages. In model C, the cross-checking method is not applied, retaining all entities and potential hallucinations in the initial answers. This results in the poorest caption generation performance. In model E, a single model BLIP-2-$\text{FlanT5}_{\mathrm{XXL}}$ is employed to verify entity words. In this approach, the entity word is retained as long as BLIP-2 considers it to exist in the image, otherwise, it is filtered out. However, the errors in the BLIP-2 model can negatively impact the accuracy of entity cross-checking, leading to sub-optimal performance. In model F, two methods are simultaneously used to filter entity words. And the entity word is retained only when both BLIP-2-$\text{FlanT5}_{\mathrm{XXL}}$ and InstructBLIP-$\text{FlanT5}_{\mathrm{XXL}}$ agree on its existence in the image. Particularly, if the judgments of two methods are contradictory, the entity word is filtered out. This strict filtering strategy is effective in reducing hallucinations, while may accidentally discard many words. Thus, although the precision reaches the highest score of 63.16\%, the recall score is notably low. 

In model G, only a larger InstructBLIP-Vicuna-13B model is used to evaluate each entity word. While this model demonstrates superior performance, errors in judgment remain inevitable. In model H and model I, the BLIP-2-$\text{FlanT5}_{\mathrm{XXL}}$ and InstructBLIP-$\text{FlanT5}_{\mathrm{XXL}}$ models are used to make preliminary judgments on entity each word. For words identified as non-existent, the InstructBLIP-Vicuna-13B model is employed for further validation. These methods adopt a more rigorous strategy, effectively reducing the risk of entity word filtering errors caused by hallucinations in a single model. By integrating InstructBLIP-Vicuna-13B to model F, the proposed HCOENet framework is established. This cross-checking method successfully eliminates hallucinations and reduces errors across multiple models, achieving the highest F1-score of 73.39\% and an average accuracy of 73.11\%.  


\subsubsection{Different Critical-Object Identification Models}


\begin{table}[!t]
	\renewcommand{\arraystretch}{1.2}
	\caption{Ablation studies of different detection methods in the critical-object identification stage. (\%)}
	\centering
	\begin{tabular}{@{}m{2.8cm}|m{0.8cm}<{\centering}m{0.5cm}<{\centering}m{0.5cm}<{\centering}m{0.9cm}<{\centering}|m{0.78cm}<{\centering}}
		\hline
		Detection Model   &Precision  &Recall  &F1-score    &Accuracy   &Average    \\
		\hline
		\cmidrule{1-6}
		None (Model A)      &\textbf{64.35} 	&80.53 	&71.53 	&67.95  &71.09         \\
		YOLOv9-E~\cite{YOLOv9}     &58.39 	&84.58 	&69.09 	&62.15    &68.55       \\     
		YOLOv10-X~\cite{YOLOv10}   &59.15 	&82.79 	&69.00 	&62.81    &68.44    \\       
		YOLOv11-X~\cite{YOLOv11}     &58.45 	&84.24 	&69.02 	&62.18    &68.47     \\    
		Tag2text~\cite{Tag2text}    &58.07 	&82.37 	&68.12 	&61.45     &67.50    \\     
		RAM~\cite{RAM}     &53.89 	&86.41 	&66.38 	&56.24   &65.73       \\     
		RAM++~\cite{RAM++}     &52.69 	&\textbf{94.15} 	&67.57 	&54.80   &67.30      \\     
		\makecell[l]{RAM++~\cite{RAM++}\&\\Grounding-DINO-B~\cite{GroundingDINO}}   &62.79 &88.30 	&\textbf{73.39} 	&\textbf{67.98} 	&\textbf{73.11}	        \\  
		\cmidrule{1-6}
	\end{tabular}
	
	\label{Table_Ablation_3}
\end{table}

In this subsection, we evaluate the effectiveness of different image detection and tagging models in the critical-object identification stage, and the results are reported in Table~\ref{Table_Ablation_3}. Model A does not contain any detection module, making it impossible to provide additional descriptions for agents that are missed in the initial response. Although this model achieves the highest precision, omitting critical targets in traffic scenarios can lead to incorrect decisions. Subsequently, we explore several different methods for object extraction. First, three state-of-the-art end-to-end detectors YOLOv9~\cite{YOLOv9}, YOLOv10~\cite{YOLOv10}, and YOLOv11~\cite{YOLOv11} are applied to detect targets in the image. These methods effectively reduce the false negatives, achieving recall improvements of 4.05\%, 2.26\%, and 3.71\%, respectively. 
Next, three advanced open-set image tagging models Tag2text~\cite{Tag2text}, RAM~\cite{RAM}, and RAM++~\cite{RAM++} are utilized to recognize all categories in the image. Among these, the RAM and RAM++ methods achieve higher recall values of 86.41\% and 94.15\%, respectively, demonstrating their superior performance in identifying objects. 

Finally, we integrate the open-set image tagging model RAM++ with the open-set detector Grounding-DINO-B to extract targets. In this method, the recognition labels generated by the RAM++ method are used as input texts for the Grounding-DINO-B model. If the Grounding-DINO-B model successfully locates the objects based on the labels, it indicates that these objects indeed exist in the image. This combined approach achieves higher accuracy and precision compared with using either the tagging or detection method independently.


\begin{table*}[!t]
	\renewcommand{\arraystretch}{1.0}
	\caption{Evaluation results of five LVLMs on the POPE benchmark under three negative sampling settings. ``Average" refers to average value of precision, recall, F1-score, and accuracy. (\%)}
	\centering
	\begin{tabular}{@{}m{2.1cm}<{\centering}m{3.0cm}|m{1.3cm}<{\centering}|m{1.3cm}<{\centering}m{1.3cm}<{\centering}m{1.4cm}<{\centering}m{1.4cm}<{\centering}|m{1.1cm}<{\centering}|m{1.4cm}<{\centering}}
		\hline
		Setting  & Model  &HCOENet    & Precision &Recall  & F1-score &Accuracy  &Average & Yes rate  \\
		\hline
		\cmidrule{1-9}
		\multirow{14}{*}{Random}   &\multirow{2}{*}{Mini‑InternVL-4B~\cite{InternVL_1.5}}  &\XSolidBrush   &92.43 	&54.89 	&68.88 	&75.20 	&72.85 	&29.69     \\
		&  &\Checkmark   &85.61 	&\textbf{77.70} 	&\textbf{81.46} 	&\textbf{82.32} 	&\textbf{81.77} 	&45.39     \\
		\cmidrule{2-9}
		& \multirow{2}{*}{LLaVA-1.5-7B~\cite{LLaVA_1.5}}  &\XSolidBrush    &83.42 	&84.00 	&83.71 	&83.65 	&83.70 	&50.35 	 \\
		&  &\Checkmark   &81.05 &\textbf{88.27} 	&\textbf{84.51} 	&\textbf{83.82} &\textbf{84.41} 		&54.46 \\
		\cmidrule{2-9}
		&  \multirow{2}{*}{mPLUG-Owl2~\cite{mplug_owl2}}  &\XSolidBrush   &85.30 	&80.59 	&82.88 	&83.35 	&83.03 	&47.24      \\
		&  &\Checkmark   &\textbf{85.86} 	&\textbf{82.63} 	&\textbf{84.22} 	&\textbf{84.51} 	&\textbf{84.31} 	&48.13   \\
		\cmidrule{2-9}
		&\multirow{2}{*}{MiniGPT-4~\cite{MiniGPT4}}  &\XSolidBrush    &62.22 	&33.28 	&43.36 	&56.54 	&48.85 	&26.74        \\
		&  &\Checkmark   &\textbf{62.73} 	&\textbf{52.10} 	&\textbf{56.92} 	&\textbf{60.57} 	&\textbf{58.08} 	&41.53      \\
		\cmidrule{2-9}
		& \multirow{2}{*}{LLaVA-1.5-13B~\cite{LLaVA_1.5}}  &\XSolidBrush   &87.65 	&81.38 	&84.40 	&84.96 	&84.60 	&46.43    \\
		&  &\Checkmark   &83.49 	&\textbf{86.64} 	&\textbf{85.03} 	&84.75    &\textbf{84.98} 		&51.89     \\
		\cmidrule{1-9}
		
		\multirow{14}{*}{Popular}   & \multirow{2}{*}{Mini‑InternVL-4B~\cite{InternVL_1.5}}  &\XSolidBrush   &74.19 	&54.79 	&63.03 	&67.86 	&64.97 	&36.93     \\
		&  &\Checkmark   &69.04 	&\textbf{77.72} 	&\textbf{73.12} 	&\textbf{71.44}   &\textbf{72.83} 		&56.28     \\
		\cmidrule{2-9}
		& \multirow{2}{*}{LLaVA-1.5-7B~\cite{LLaVA_1.5}}  &\XSolidBrush   & 59.82 	&83.96 	&69.87 	&63.79 	&69.36 	&70.17   \\
		&  &\Checkmark   &\textbf{62.79} &\textbf{88.30} 	&\textbf{73.39} 	&\textbf{67.98} 	&\textbf{73.12}	 &70.32    \\
		\cmidrule{2-9}
		&  \multirow{2}{*}{mPLUG-Owl2~\cite{mplug_owl2}}  &\XSolidBrush   &58.90 	&81.02 	&68.21 	&62.25 	&67.60 	&68.78     \\
		&  &\Checkmark   &\textbf{61.64} 	&\textbf{83.54} 	&\textbf{70.94} 	&\textbf{65.77}   &\textbf{70.47} 	  &67.77     \\
		\cmidrule{2-9}
		&\multirow{2}{*}{MiniGPT-4~\cite{MiniGPT4}}  &\XSolidBrush   &57.91 	&32.24 	&41.41 	&54.40 	&46.49 	&27.83        \\
		&  &\Checkmark   &\textbf{57.98} 	&\textbf{52.16} 	&\textbf{54.91} 	&\textbf{57.18} 	&\textbf{55.56} 	&44.98       \\
		\cmidrule{2-9}
		& \multirow{2}{*}{LLaVA-1.5-13B~\cite{LLaVA_1.5}}  &\XSolidBrush   &63.19 	&81.33 	&71.12 	&66.98 	&70.66 	&64.35    \\
		&  &\Checkmark   &\textbf{64.89} &\textbf{86.66} 	&\textbf{74.21} 	&\textbf{69.88} 	&\textbf{73.91} 	&66.78   \\
		\cmidrule{1-9}
		
		\multirow{14}{*}{Adversarial}   &  \multirow{2}{*}{Mini‑InternVL-4B~\cite{InternVL_1.5}}  &\XSolidBrush   &75.12	&54.79 	&63.37 	&68.32 	&65.40 	&36.48       \\
		&  &\Checkmark   &69.83 	&\textbf{77.71} 	&\textbf{73.56} 	&\textbf{72.07} &\textbf{73.29} 		&55.65     \\
		\cmidrule{2-9}
		& \multirow{2}{*}{LLaVA-1.5-7B~\cite{LLaVA_1.5}}  &\XSolidBrush  &61.64 	&83.96 	&71.09 	&65.85 	&70.64 	&68.11   \\
		&  &\Checkmark   &\textbf{63.89}  &\textbf{88.34} 	&\textbf{74.14} 	&\textbf{69.20} 	&\textbf{73.89} 	&69.14 \\
		\cmidrule{2-9}
		&  \multirow{2}{*}{mPLUG-Owl2~\cite{mplug_owl2}}  &\XSolidBrush   &60.53 	&81.15 	&69.34 	&64.11 	&68.78 	&67.03     \\
		&  &\Checkmark   &\textbf{63.37} 	&\textbf{83.57} 	&\textbf{72.08} 	&\textbf{67.63} 	&\textbf{71.66} 	&65.94      \\
		\cmidrule{2-9}
		&\multirow{2}{*}{MiniGPT-4~\cite{MiniGPT4}}  &\XSolidBrush   &57.98 	&31.83 	&41.10 	&54.38 	&46.32 	&27.45        \\
		&  &\Checkmark   &57.69 	&\textbf{51.87} 	&\textbf{54.63} 	&\textbf{56.91} 	&\textbf{55.27} 	&44.96     \\
		\cmidrule{2-9}
		& \multirow{2}{*}{LLaVA-1.5-13B~\cite{LLaVA_1.5}}  &\XSolidBrush   &64.95 	&81.13 	&72.14 	&68.67 	&71.72 	&62.46   \\
		&  &\Checkmark   &\textbf{66.13} 	&\textbf{86.67} 	&\textbf{75.02} 	&\textbf{71.13}   &\textbf{74.74} 		&65.54     \\
		\cmidrule{1-9}
	\end{tabular}
	
	\label{Table_POPE_VLM_results}
\end{table*}

\subsection{Results on POPE Benchmark}
\subsubsection{Results of Multiple LVLMs}

In this subsection, we conduct experiments on five LVLMs to evaluate the effectiveness of HCOENet in alleviating object hallucinations. All results are obtained on the POPE benchmark under three sampling settings, which are reported in Table~\ref{Table_POPE_VLM_results}. First, MiniGPT-4 model demonstrates the weakest vision-language comprehension ability, producing the highest number of hallucinations under the random sampling, with an F1-score of only 43.36\% and the model accuracy of 56.54\%. After incorporating the HCOENet framework, the F1-score and accuracy increase by 13.56\% and 4.03\%, respectively. Mini-InternVL-4B, with only 4 billion parameters, has the fewest parameters among the LVLMs. Despite this, applying HCOENet yields significant improvements in object-level hallucination suppression, with an F1-score increase of 12.58\% and an accuracy increase of 7.12\%. Additionally, LLaVA-1.5-13B has the largest number of parameters and the strongest perceptual ability, while still exhibits hallucinations. This finding emphasizes that merely increasing the size of LVLMs is insufficient to fully eliminate hallucinations, highlighting the necessity of using the HCOENet framework to address this issue. 

In the more challenging popular and adversarial settings, the performance of most LVLMs significantly decreases. To be exact, the F1-score of LLaVA-1.5-7B decreases from 83.71\% to 69.87\%, while the F1-score of mPLUG-Owl2 declines from 82.88\% to 68.21\%. Nonetheless, the HCOENet still remains effective, achieving substantial improvements across various metrics compared with the baselines. For example, Mini-InternVL-4B model shows an average improvement of 7.86\%, while MiniGPT-4 model achieves an average improvement of 9.07\%. Moreover, the results reveal that Mini-InternVL-4B and MiniGPT-4 exhibit lack of confidence excessively, reflected in answering \textit{no} for many questions and having a lower Yes rate. This phenomenon is also mitigated by applying our method. MiniGPT-4, in particular, shows the greatest improvement in model confidence, with an increase of over 17\% in the Yes rate under the popular and adversarial setting.


\subsubsection{Comparison with State-of-the-art Methods}

\begin{table*}[!t]
	\renewcommand{\arraystretch}{1.2}
	\caption{Comparison results on the POPE benchmark. ``Average" refers to average value of precision, recall, F1-score, and accuracy. (\%)}
	\centering
	\begin{tabular}{@{}m{2.3cm}<{\centering}|m{2.4cm}|m{0.9cm}<{\centering}m{1.1cm}<{\centering}m{0.9cm}<{\centering}|m{0.9cm}<{\centering}m{1.1cm}<{\centering}m{0.9cm}<{\centering}|m{0.9cm}<{\centering}m{1.1cm}<{\centering}m{0.9cm}<{\centering}}
		\hline
		\multirow{2}{*}{Model}  &\multirow{2}{*}{Method}   & \multicolumn{3}{c|}{Random}  & \multicolumn{3}{c|}{Popular}  & \multicolumn{3}{c}{Adversarial} \\
		&   &Recall  &F1-score   &Average  &Recall  &F1-score &Average  &Recall &F1-score    &Average        \\
		\hline
		\cmidrule{1-11}
		\multirow{6}{*}{LLaVA-1.5-7B~\cite{LLaVA_1.5}}  &vanilla  &84.12 	 &83.60 	&83.58   &84.08    &69.87 	 &69.37    &84.05 	&71.11 	 &70.66       \\
		&Woodpecker~\cite{woodpecker}  &77.69  &80.93 	 &81.19      &77.53   &71.33 	&70.93      &77.51  &72.50 	 &72.17        \\
		&ReCaption~\cite{Recaption}  &83.62  &83.91  &83.90    &84.24   &70.04 		&69.55    &84.27    &71.29	  &70.84        \\
		&VCD~\cite{VCD}   &86.57  &79.83  &79.65      &86.52 	&70.10 	 &69.65  	&86.49    &71.02  &70.62        \\
		&Ours (HCNet)   &80.50   &82.61  &82.75       &80.53  	&71.53  	 &71.09   	&80.50     &72.67     &72.28      \\
		&Ours (HCOENet)   &\textbf{88.27}  &\textbf{84.51}  &\textbf{84.41}    &\textbf{88.30}    &\textbf{73.39}    &\textbf{73.11}    	&\textbf{88.34}   	&\textbf{74.14}    &\textbf{73.89}       \\
		\cmidrule{1-11}
		\addlinespace[-0.6ex]
		
		\multirow{6}{*}{LLaVA-1.5-13B~\cite{LLaVA_1.5}}  &vanilla  &81.17 	 &84.18 	&84.38   	&81.05   &70.93   &70.46   &81.20    &72.19   &71.77       \\ 
		&Woodpecker~\cite{woodpecker} 	&77.69  &80.93  	&81.19   &75.28     &72.27   &72.04      &75.22    &73.04 	 &72.86         \\
		&ReCaption~\cite{Recaption}   &81.64   &84.31 	&84.48    	&81.62   &71.39  	 &70.94       &81.57  &72.53 	  &72.12        \\
		&VCD~\cite{VCD}    &83.98  &81.21 	 &81.10      &83.93   &70.68 	&70.21    &83.74 	     &71.37 	 &70.92        \\
		&Ours (HCNet)   &76.93 	&82.44    &82.95    &77.02 	&72.45 
		&72.14    &76.86 	&73.37 
		&73.13   \\
		&Ours (HCOENet)   &\textbf{86.64}  &\textbf{85.03}   &\textbf{84.98}     &\textbf{86.66}  &\textbf{74.21} 	&\textbf{73.91}     &\textbf{86.67} 	&\textbf{75.02}   &\textbf{74.73}       \\
		\cmidrule{1-11}
		\addlinespace[-0.6ex]
		
		\multirow{8}{*}{MiniGPT-4~\cite{MiniGPT4}}  &vanilla   &33.28 	 &43.36	   &48.85   	&32.24   &41.41   &46.49  &31.83    &41.10  &46.32       \\ 
		&LRV-Instruction~\cite{LRV_Instruction}  &42.68 	&51.07  	&54.10 &42.74 	&48.61 	&50.63  &42.70 	&48.81 	&50.93       \\ 
		&HalluciDoctor (w/LLaVA+)~\cite{hallucidoctor}  &26.37 	&39.79 	&51.82  &26.35 	&37.13 	&45.43 &29.09 	&39.95 &47.27      \\ 
		&HalluciDoctor (w/LLaVA++)~\cite{hallucidoctor} &44.34 	&56.00 	&\textbf{60.36} &44.51 	&50.94 	&53.03   &44.57 	&51.61 	&53.91     \\ 
		&Ours (HCNet)   &24.87 	&34.65  &42.42     &24.98 	&34.54 
		&42.02   	 &24.82 	&34.36     &41.88      \\
		&Ours (HCOENet)  &\textbf{52.10} 	&\textbf{56.92} 	&58.08 &\textbf{52.16} 	&\textbf{54.91} 	&\textbf{55.56}   &\textbf{51.87}    &\textbf{54.63}  &\textbf{55.27}       \\ 
		\cmidrule{1-11}
	\end{tabular}
	
	\label{Table_comparison_results}
\end{table*}

\begin{figure}[!t]
	\centering
	\includegraphics[width=1.0\linewidth]{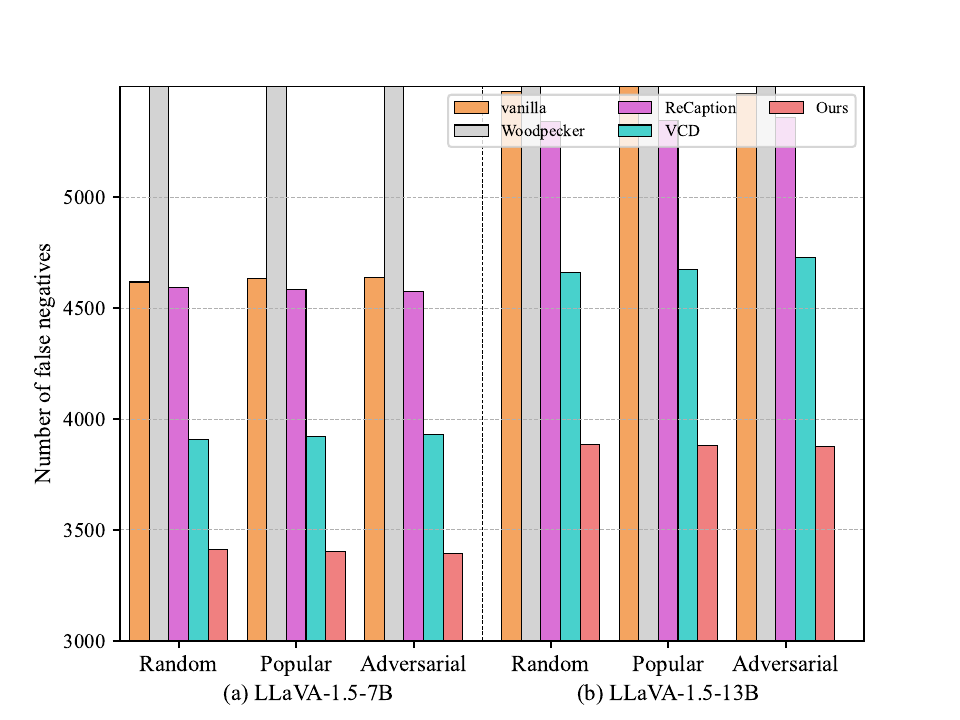}
	\caption{The number of false negatives on the LLaVA-1.5-7B and LLaVA-1.5-13B models under three negative sampling settings.}
	\label{Fig_FN_samples}
\end{figure}

In this subsection, the proposed HCOENet is first compared with Woodpecker~\cite{woodpecker}, ReCaption~\cite{Recaption}, and VCD~\cite{VCD} on the LLaVA-1.5-7B and LLaVA-1.5-13B model. The comparison results, presented in Table~\ref{Table_comparison_results}, indicate that the HCOENet achieves significant improvements in alleviating object hallucinations, particularly in the recall and F1-score metric. Notably, the increase in recall is especially prominent, for instance, HCOENet surpasses the Woodpecker model by more than 10\% and the ReCaption model by approximately 5\%. In general, a higher recall value represents fewer false negatives, signifying that fewer traffic participants in the scene are mistakenly identified as non-existent. This is critical for ensuring accurate driving decisions in traffic scenarios. Here, Fig.~\ref{Fig_FN_samples} compares the number of false negatives across different hallucination mitigation methods. Except for the Woodpecker method, the other three models effectively reduce the number of false negatives when compared with the vanilla, with HCOENet minimizing false negatives. 

Next, the proposed HCNet and HCOENet are compared with LRV-Instruction~\cite{LRV_Instruction} and HalluciDoctor~\cite{hallucidoctor} based on the MiniGPT-4 model. Unlike previous methods, which primarily focus on suppressing hallucinatory information in the texts generated by LVLMs, these approaches struggle to describe omitted objects, resulting in modest improvements in recall values. In contrast, the critical-object identification and description modules in the HCOENet effectively enrich semantic information, greatly improving the recall value. Compared with the HalluciDoctor model fine-tuned on the LLaVA++ dataset, our method adopts a post-processing architecture that achieves state-of-the-art results without requiring any fine-tuning on additional datasets. This significantly reduces training costs while delivering superior performance.


\subsubsection{Comparison between Different mPLUG-Owl Models}

\begin{table}[!t]
	\renewcommand{\arraystretch}{1.0}
	\caption{Comparison results between different mPLUG-Owl models on the POPE benchmark. ``w/Ours" denotes using HCOENet to correct the response generated by LVLMs. (\%)}
	\centering
	\begin{tabular}{@{}m{1.15cm}<{\centering}m{2.8cm}|m{1.6cm}m{1.7cm}}
		\hline
		Setting    &Model   &F1-score   &Accuracy    \\
		\hline
		\cmidrule{1-4}
		\multirow{8}{*}{Random}  &mPLUG-Owl~\cite{mplug_owl}    &65.42 	&56.24   \\  
		&mPLUG-Owl w/Ours     &67.81 ($\uparrow$\textbf{2.39})   &58.89  ($\uparrow$\textbf{2.65}) \\  
		&mPLUG-Owl2~\cite{mplug_owl2}     &82.88    &83.35    \\
		&mPLUG-Owl2 w/Ours    &84.22 ($\uparrow$\textbf{1.34})    &84.51 ($\uparrow$\textbf{1.16})     \\
		&mPLUG-Owl2.1~\cite{mplug_owl2}    &81.51     &82.69     \\
		&mPLUG-Owl2.1 w/Ours   &83.39 ($\uparrow$\textbf{1.88})    &84.07 ($\uparrow$\textbf{1.38})    \\
		&mPLUG-Owl3~\cite{mplug_owl3}    &85.73 	&86.19          \\
		&mPLUG-Owl3 w/Ours    &86.35 ($\uparrow$\textbf{0.62})	 &86.88 ($\uparrow$\textbf{0.69})     \\
		\cmidrule{1-4}
		\multirow{8}{*}{Popular}   &mPLUG-Owl~\cite{mplug_owl}   &60.45 &51.97  \\  
		&mPLUG-Owl w/Ours     &65.28 ($\uparrow$\textbf{4.83})   & 54.12 ($\uparrow$\textbf{2.15})   \\  
		& mPLUG-Owl2~\cite{mplug_owl2}   &68.21    &62.25     \\
		& mPLUG-Owl2 w/Ours   &70.94 ($\uparrow$\textbf{2.73})   &65.77 ($\uparrow$\textbf{3.52})    \\
		&mPLUG-Owl2.1~\cite{mplug_owl2}    &67.85    &63.32    \\
		&mPLUG-Owl2.1 w/Ours    &72.79 ($\uparrow$\textbf{4.94})   &70.10 ($\uparrow$\textbf{6.78})   \\
		&mPLUG-Owl3~\cite{mplug_owl3} 	&73.37 	 &70.34 	 	\\
		&mPLUG-Owl3 w/Ours	 &77.65 ($\uparrow$\textbf{4.28})  	&76.12 ($\uparrow$\textbf{5.78})   	\\
		\cmidrule{1-4}
		\multirow{8}{*}{Adversarial}   &mPLUG-Owl~\cite{mplug_owl}    &60.54 	&52.38     \\  
		&mPLUG-Owl w/Ours     & 65.72 ($\uparrow$\textbf{5.18})  &   54.83 ($\uparrow$\textbf{2.45})   \\  
		& mPLUG-Owl2~\cite{mplug_owl2}    &69.34    &64.11     \\
		& mPLUG-Owl2 w/Ours    &72.08 ($\uparrow$\textbf{2.74})   &67.63 ($\uparrow$\textbf{3.52})    \\
		&mPLUG-Owl2.1~\cite{mplug_owl2}    &68.84    &63.57  \\
		&mPLUG-Owl2.1 w/Ours    &73.75 ($\uparrow$\textbf{4.91})   &71.54 ($\uparrow$\textbf{7.97})  \\
		&mPLUG-Owl3~\cite{mplug_owl3}     &74.69 	&71.89      \\
		&mPLUG-Owl3 w/Ours    &78.80 ($\uparrow$\textbf{4.11})	 &77.68 ($\uparrow$\textbf{5.79})       \\
		\cmidrule{1-4}
	\end{tabular}
	\label{Table_compare_three_mplug_owl_models}
\end{table}

In Table~\ref{Table_compare_three_mplug_owl_models}, four versions of mPLUG-Owl models (mPLUG-Owl, mPLUG-Owl2, mPLUG-Owl2.1, mPLUG-Owl3) released in different years are compared under three negative sampling settings, with evaluation metrics of F1-score and model accuracy. The results highlight the primary differences between four models, which stem from variations in their visual encoders and language decoders. Among these, mPLUG-Owl3 leverages the advanced visual encoder SigLIP-400M and the language generator Qwen2, achieving the highest accuracy and the lowest hallucination levels across all negative sampling methods. Concretely, it attains an F1-score of 85.73\% and an accuracy of 86.19\% in the random setting. In contrast, mPLUG-Owl model exhibits the weakest scene comprehension ability, generating descriptions with significant inconsistencies relative to the given image. Even under the simplest random sampling method, its F1-score and accuracy are only 65.42\% and 56.24\%, respectively. The mPLUG-Owl2 and mPLUG-Owl2.1 models demonstrate relatively robust visual understanding ability, providing more comprehensive interpretations of traffic scenarios.

Subsequently, the proposed HCOENet is applied to each mPLUG-Owl model, yielding significant improvements in each evaluation metric under three different sampling settings. Among them, mPLUG-Owl obtains the most substantial enhancement under the random sampling, with the F1-score and accuracy increasing by 2.39\% and 2.65\%, respectively. Under the popular sampling setting, mPLUG-Owl2.1 exhibits the greatest improvement. with the F1-score and model accuracy increasing by 4.94\% and 6.78\%, respectively. These results highlight the strong generalization capability of HCOENet, demonstrating its effectiveness in eliminating semantic hallucinations across various LVLMs.

\subsubsection{Comparison between Different InternVL Models}

\begin{table}[!t]
	\renewcommand{\arraystretch}{1.2}
	\caption{Comparison results between different InternVL models on the POPE benchmark. ``w/Ours" denotes using HCOENet to correct the response generated by LVLMs. (\%)}
	\centering
	\begin{tabular}{@{}m{1.2cm}<{\centering}m{3.1cm}|m{0.8cm}|m{0.7cm}m{1.0cm}}
		\hline
		Setting    &Model   &Params   &Recall  &F1-score     \\
		\hline
		\cmidrule{1-5}
		\multirow{5}{*}{Random} &Mini-InternVL-4B~\cite{InternVL_1.5}   &4.2B     &54.89 	&68.88         \\
		&InternVL2-4B~\cite{InternVL_1.5}   &4.2B     &69.77 	&79.64        \\
		&InternVL2-8B~\cite{InternVL_1.5}   &8B     &69.76 	&79.43         \\
		&InternVL2-40B~\cite{InternVL_1.5}   &40B     &68.62 	&79.56 
		\\
		&Mini-InternVL-4B w/Ours   &47B    &\textbf{77.70}   &\textbf{81.46}         \\
		\cmidrule{1-5}
		\multirow{5}{*}{Popular} &Mini-InternVL-4B~\cite{InternVL_1.5}   &4.2B     &54.79 	&63.03         \\
		&InternVL2-4B~\cite{InternVL_1.5}   &4.2B     &69.77 	&72.84       \\
		&InternVL2-8B~\cite{InternVL_1.5}   &8B     &69.76 	&72.64      \\
		&InternVL2-40B~\cite{InternVL_1.5}   &40B     &68.62 	&\textbf{75.24}    \\
		& Mini-InternVL-4B w/Ours   &47B   &\textbf{77.72}   &73.12     \\
		\cmidrule{1-5}
		\multirow{5}{*}{Adversarial} &Mini-InternVL-4B~\cite{InternVL_1.5}   &4.2B     &54.79 	&63.37         \\
		&InternVL2-4B~\cite{InternVL_1.5}   &4.2B     &69.77 	&73.46       \\
		&InternVL2-8B~\cite{InternVL_1.5}   &8B     &69.76 	&73.40       \\
		&InternVL2-40B~\cite{InternVL_1.5}      &40B     &68.62 	&\textbf{75.89}     \\
		& Mini-InternVL-4B w/Ours       &47B      &\textbf{77.71}   &73.56      \\
		\cmidrule{1-5}
	\end{tabular}
	
	\label{Table_compare_with_internvl2}
\end{table}

In Table~\ref{Table_compare_with_internvl2}, we compare the performance of Mini-InternVL-4B model with more advanced InternVL2-4B, InternVL2-8B, and InternVL2-40B model. Due to the more powerful visual encoder and language generator in the InternVL2-40B model, it has superior visual question-answering capabilities. Although the performance of the vanilla Mini-InternVL-4B model is much lower than the InternVL2-8B model, it surpasses the latter after applying chain-of-thought corrections through the HCOENet. This result underscores the effectiveness of the proposed architecture in eliminating hallucinations. In terms of model size, both the Mini-InternVL-4B method with HCOENet and the InternVL2-40B method have substantial parameters exceeding 40 billion. Under the random sampling setting, the smaller Mini-InternVL-4B method with HCOENet outperforms the InternVL2-40B method, achieving a 9.08\% increase in the recall value and a 1.9\% increase in the F1-score. In the more complex popular and adversarial settings, the Mini-InternVL-4B with HCOENet maintains strong semantic understanding ability, significantly reducing the false negatives and delivering F1-scores comparable to those of the larger InternVL2-40B model.



\begin{figure}[!t]
	\centering
	\includegraphics[width=1.0\linewidth]{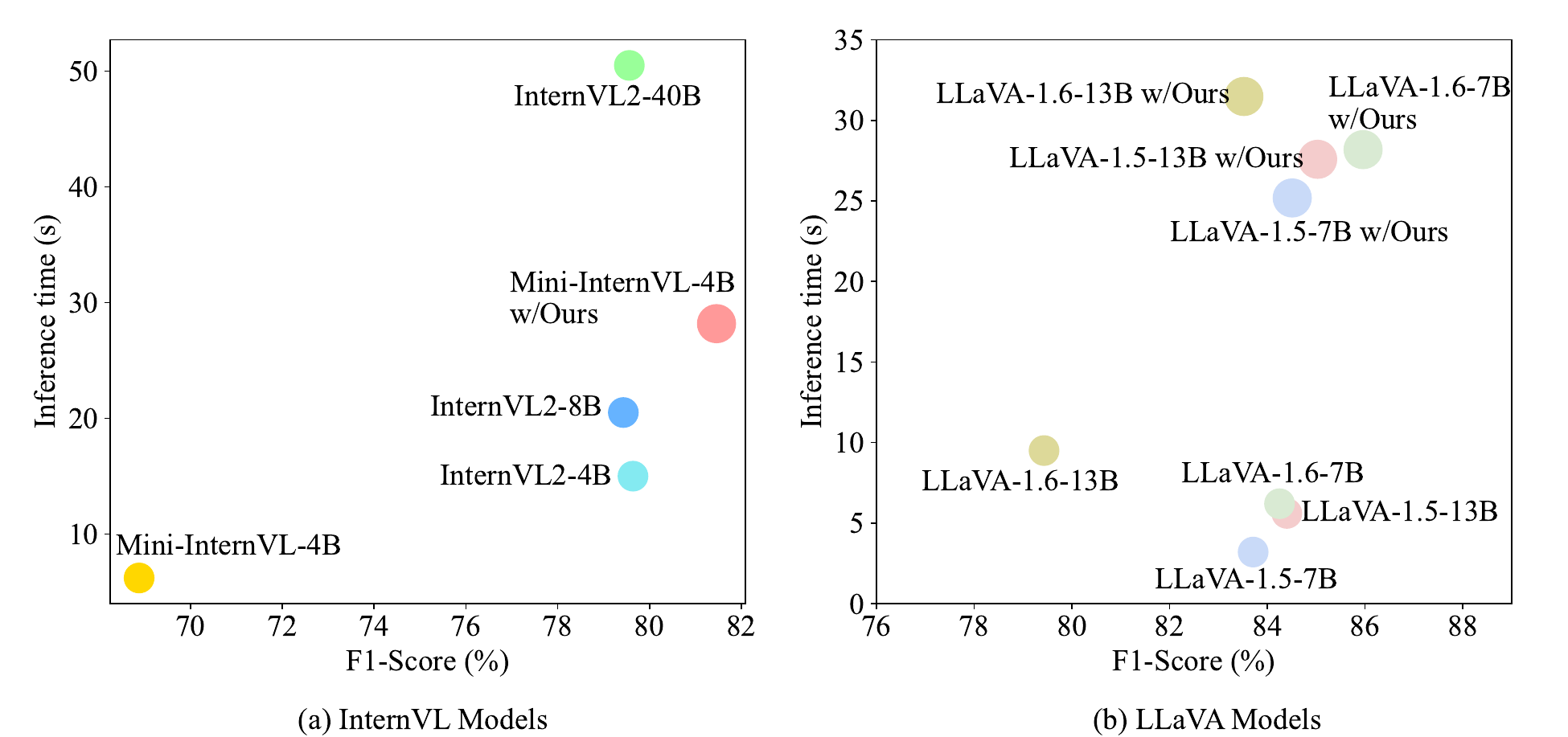}
	\caption{Inference time and F1-score of different InternVL models and LLaVA models. (The F1-score is computed under the random sampling setting on the POPE benchmark.)}
	\label{Fig_internvl_and_llava}
\end{figure}

Furthermore, Fig.~\ref{Fig_internvl_and_llava}(a) compares the inference time and F1-score values of different InternVL models. The lightweight Mini-InternVL-4B method requires the shortest inference time, generating descriptions for an image in just 6.2 seconds. In contrast, the larger InternVL2-8B and InternVL2-40B models, due to their greater number of parameters, require considerably more time for inference, taking 20.5 seconds and 50.5 seconds, respectively. When using the Mini-InternVL-4B model with HCOENet, the inference time increases to 28.2 seconds, still approximately 44\% faster than the InternVL2-40B model used alone. The above results highlight the advantages of using the proposed HCOENet for eliminating hallucinations in small-scale LVLMs. The proposed framework not only improves model performance but also achieves a better balance between performance and inference time, providing faster inference and comparable results to those of large-scale models.

\subsubsection{Comparison between Different LLaVA Models}

\begin{table}[!t]
	\renewcommand{\arraystretch}{1.0}
	\caption{Comparison results between different LLaVA models on the POPE benchmark. ``w/Ours" denotes using HCOENet to correct the response generated by LVLMs. (\%)}
	\centering
	\begin{tabular}{@{}m{1.2cm}<{\centering}m{2.9cm}|m{1.6cm}m{1.6cm}}
		\hline
		Setting    &Model    &Recall     &F1-score    \\
		\hline
		\cmidrule{1-4}
		\multirow{8}{*}{Random}  &LLaVA-1.5-7B~\cite{LLaVA_1.5}   &84.00 &83.71           \\
		&LLaVA-1.5-7B w/Ours   &88.27 ($\uparrow$\textbf{4.27})       &84.51 ($\uparrow$\textbf{0.80})         \\
		&LLaVA-1.5-13B~\cite{LLaVA_1.5}    &81.38         &84.40       \\
		&LLaVA-1.5-13B w/Ours   &86.64 ($\uparrow$\textbf{5.26})       &85.03 ($\uparrow$\textbf{0.63})        \\
		&LLaVA-1.6-7B~\cite{LLaVA_1.6}    &80.94       &84.25       \\
		&LLaVA-1.6-7B w/Ours     &86.46 ($\uparrow$\textbf{5.52})      &85.96 ($\uparrow$\textbf{1.71})	   \\
		&LLaVA-1.6-13B~\cite{LLaVA_1.6}     &70.65       &79.43        \\
		&LLaVA-1.6-13B w/Ours    &79.12 ($\uparrow$\textbf{8.47})        &83.52 ($\uparrow$\textbf{4.09})	  \\
		\cmidrule{1-4}
		\multirow{8}{*}{Popular}  &LLaVA-1.5-7B~\cite{LLaVA_1.5}  &83.96        &69.87          \\
		&LLaVA-1.5-7B w/Ours    &88.30 ($\uparrow$\textbf{4.34})      &73.39 ($\uparrow$\textbf{3.52})          \\
		&LLaVA-1.5-13B~\cite{LLaVA_1.5}    &81.33         &71.12        \\
		&LLaVA-1.5-13B w/Ours  &86.66 ($\uparrow$\textbf{5.33})   &74.21 ($\uparrow$\textbf{3.09})         \\
		&LLaVA-1.6-7B~\cite{LLaVA_1.6}    &80.85         &73.93         \\
		&LLaVA-1.6-7B w/Ours     &86.30 ($\uparrow$\textbf{5.45})        &74.89 ($\uparrow$\textbf{0.96})	    \\
		&LLaVA-1.6-13B~\cite{LLaVA_1.6}    &70.59        &70.29     \\
		&LLaVA-1.6-13B w/Ours    &79.01 ($\uparrow$\textbf{8.42})        &73.36 ($\uparrow$\textbf{3.07})	       \\
		\cmidrule{1-4}
		\multirow{8}{*}{Adversarial}  &LLaVA-1.5-7B~\cite{LLaVA_1.5}  &83.96        &71.09          \\
		&LLaVA-1.5-7B w/Ours    &88.34 ($\uparrow$\textbf{4.38})      &74.14 ($\uparrow$\textbf{3.05})            \\
		&LLaVA-1.5-13B~\cite{LLaVA_1.5}    &81.13       &72.14            \\
		&LLaVA-1.5-13B w/Ours    &86.67 ($\uparrow$\textbf{5.54})      &75.02 ($\uparrow$\textbf{2.88})         \\
		&LLaVA-1.6-7B~\cite{LLaVA_1.6}   &80.94     &75.00    \\
		&LLaVA-1.6-7B w/Ours   &86.37 ($\uparrow$\textbf{4.43})   &76.09 ($\uparrow$\textbf{1.09})  	     \\
		&LLaVA-1.6-13B~\cite{LLaVA_1.6}   &70.69   &71.18      \\
		&LLaVA-1.6-13B w/Ours    &79.29 ($\uparrow$\textbf{8.60})   &74.60 ($\uparrow$\textbf{3.42})	      \\
		\cmidrule{1-4}
		
	\end{tabular}
	
	\label{Table_compare_with_llava}
\end{table}

In Table~\ref{Table_compare_with_llava}, we compare the LLaVA-1.5-7B and LLaVA-1.5-13B model with the more advanced LLaVA-1.6-7B and LLaVA-1.6-13B model. The larger language decoder in the LLaVA-1.5-13B model enhances its scene description capabilities, outperforming the LLaVA-1.5-7B model by 0.69\% and 1.25\% in the F1-score under the random and popular samplings, respectively. Among the four multimodal methods, LLaVA-1.5-7B model obtains the highest recall score and generates the fewest false negatives. Meanwhile, LLaVA-1.6-7B model demonstrates the best comprehensive understanding of traffic scenarios, achieving F1-scores of 73.93\% and 75.0\% under the popular and adversarial sampling settings, respectively.   

However, the LLaVA-1.6-13B model, despite containing the most parameters, performs poorly in traffic scenarios, generating numerous hallucinations and achieving only an F1-score of 71.18\% under the adversarial sampling setting. After integrating the HCOENet, the F1-score shows significant improvements, increasing by 4.09\% and 3.42\% in the random and adversarial settings, respectively. Fig.~\ref{Fig_internvl_and_llava}(b) compares the inference time and F1-score of different LLaVA models. Compared with models with 7 billion parameters, the LLaVA models with 13 billion parameters require approximately 3 seconds longer to process an image. Additionally, adopting HCOENet increases the inference time by an average of about 20 seconds but substantially improves the performance of the multimodal models.

\subsection{Result Analysis}
\subsubsection{Comparison with GPT-4o Model} 

\begin{table}[!t]
	\renewcommand{\arraystretch}{1.2}
	\caption{Comparison with the GPT-4o model on the POPE benchmark. ``w/Ours" denotes using HCOENet to correct the response generated by LVLMs, ``B" denotes billion, and ``T" denotes trillion. (\%)}
	\centering
	\begin{tabular}{@{}m{1.1cm}<{\centering}m{2.75cm}|m{0.9cm}|m{1.15cm}m{1.0cm}}
		\hline
		Setting    &Model   &Params$\downarrow$   &F1-score$\uparrow$  &Accuracy$\uparrow$   \\
		\hline
		\cmidrule{1-5}
		\multirow{6}{*}{Random} 
		&LLaVA-1.5-7B w/Ours   &50B      &84.05   &83.08   \\
		&LLaVA-1.5-13B w/Ours   &56B      &85.05  &84.50    \\
		&LLaVA-1.6-7B w/Ours   &50B      &86.04  &85.83    \\
		&mPLUG-Owl2 w/Ours   &51B      &84.31    &83.75  \\
		&mPLUG-Owl3 w/Ours   &51B      &\textbf{87.38}    &\textbf{87.75}  \\
		&GPT-4o~\cite{GPT_4}   &$\sim$1.8T    &86.77   &86.58         \\
		\cmidrule{1-5}
		\multirow{6}{*}{Popular} 
		&LLaVA-1.5-7B w/Ours   &50B    &73.43 	&67.92    \\
		&LLaVA-1.5-13B w/Ours   &56B    &74.23 	&69.50   \\
		&LLaVA-1.6-7B w/Ours   &50B    &73.47 	&67.92    \\
		&mPLUG-Owl2 w/Ours   &51B    &72.13 	&66.00   \\
		&mPLUG-Owl3 w/Ours   &51B    &\textbf{78.91} 	&\textbf{77.33}    \\
		&GPT-4o~\cite{GPT_4}   &$\sim$1.8T    &77.97   &75.42         \\
		\cmidrule{1-5}
		\multirow{6}{*}{Adversarial} 
		&LLaVA-1.5-7B w/Ours   &50B    &74.13 	&69.00    \\
		&LLaVA-1.5-13B w/Ours   &56B    &74.95 	&70.58     \\
		&LLaVA-1.6-7B w/Ours   &50B    &76.88 	&73.83    \\
		&mPLUG-Owl2 w/Ours   &51B    &72.45 	&68.00      \\
		&mPLUG-Owl3 w/Ours   &51B    &\textbf{80.28} 	&\textbf{79.17}    \\
		&GPT-4o~\cite{GPT_4}   &$\sim$1.8T    &79.70 	&77.67   \\
		\cmidrule{1-5}
	\end{tabular}
	
	\label{Table_compare_with_GPT}
\end{table}

\begin{figure}[!t]
	\centering
	\includegraphics[width=1.0\linewidth]{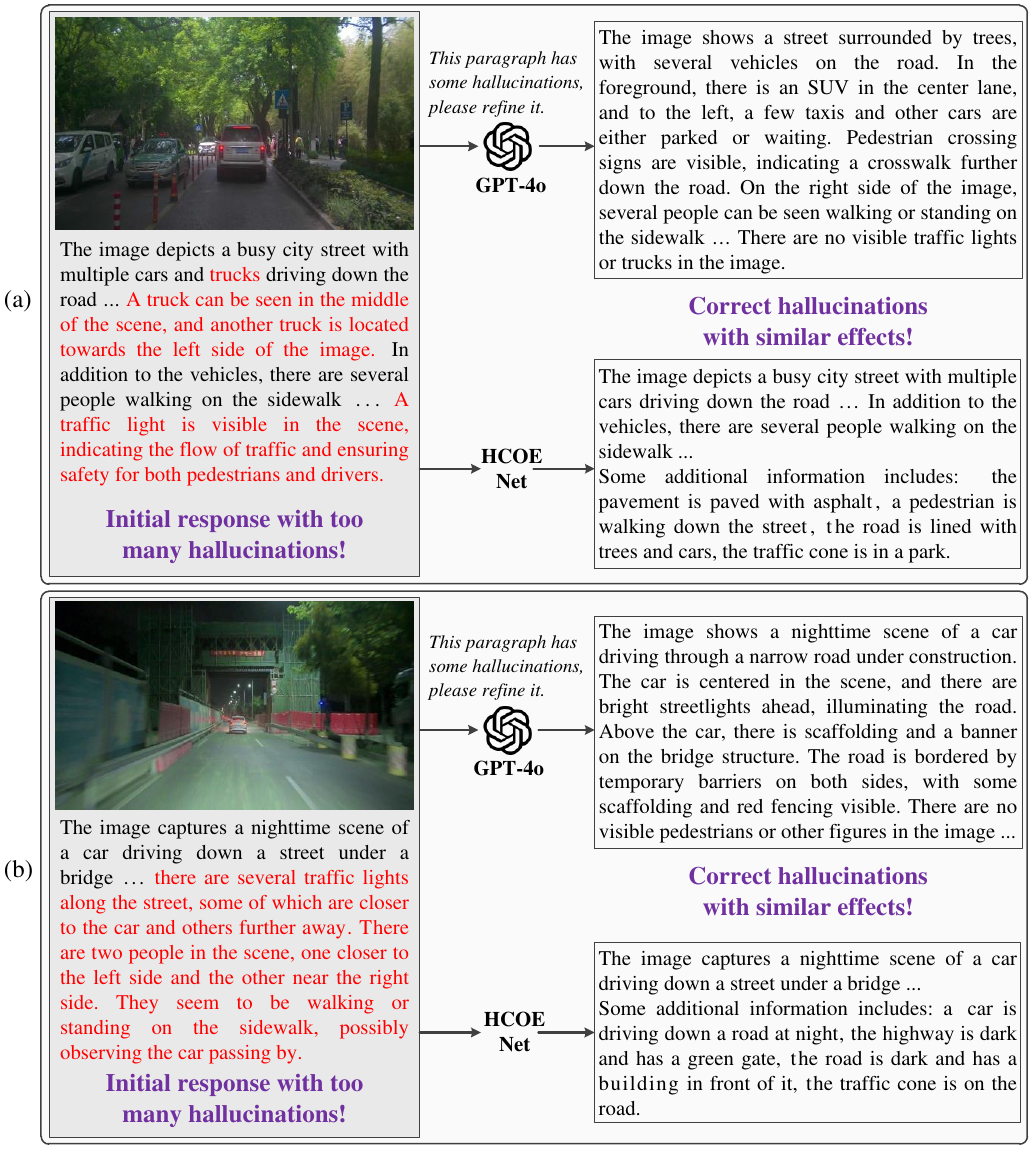}
	\caption{Hallucination elimination results in comparison with the GPT-4o model, with hallucinatory texts highlighted in red.}
	\label{Fig_compare_gpt4o}
\end{figure}

In Table~\ref{Table_compare_with_GPT}, we compare the LLaVA-1.5-7B with HCOENet, LLaVA-1.5-13B model with HCOENet, LLaVA-1.6-7B with HCOENet, mPLUG-Owl2 with HCOENet, mPLUG-Owl3 model with HCOENet, and the state-of-the-art GPT-4o~\cite{GPT_4} model on the POPE benchmark. And only 200 randomly selected images are utilized to calculate the F1-score and model accuracy metric. These results show the descriptions generated by our framework and those directly produced using GPT-4o have a similar level of object hallucinations. For example, the mPLUG-Owl3 model with HCOENet achieves an F1-score of 87.38\%, while the GPT-4o model achieves a similar score of 86.77\% under the random sampling setting. Therefore, the LVLM with HCOENet serves as an alternative to the GPT-4o model for image description, providing comparable text generation capabilities while significantly reducing token costs. In terms of model size, HCOENet contains approximately 43 billion parameters, while the GPT-4o model has over 1.8 trillion parameters, about 36 times larger than our model. 


Moreover, we qualitatively compare the performance of HCOENet and GPT-4o method in Fig.~\ref{Fig_compare_gpt4o}. The types and numbers of traffic participants in Fig.~\ref{Fig_compare_gpt4o}(a) are quite complex, leading to a significant number of hallucinations in the initial response generated by the LVLM. GPT-4o is able to accurately identify the erroneous information in the paragraph and refine the captions, such as removing false mentions of traffic lights or trucks that are not visible in the image. Similarly, HCOENet successfully identifies and removes these hallucinatory elements. In Fig.~\ref{Fig_compare_gpt4o}(b), the blurred perspective poses a significant challenge to the visual feature extraction ability for LVLMs, resulting in over half of the predicted response being incorrect. For instance, streetlights are misidentified as traffic lights, and although there are no people in the image, the model incorrectly describes pedestrians as \textit{``walking or standing on the sidewalk"}. After correction by GPT-4o, the refined texts precisely indicate the absence of visible pedestrians and correctly describe bright headlights ahead in the scene. In comparison, HCOENet not only removes the hallucinations related to traffic lights and pedestrians but also goes a step further by identifying additional objects, such as traffic cones on the road. 
	
The qualitative and quantitative results above demonstrate that HCOENet achieves image description and semantic hallucination mitigation performance comparable to the state-of-the-art GPT-4o model.


\subsubsection{Inference Time}

\begin{figure}[!t]
	\centering
	\includegraphics[width=0.9\linewidth]{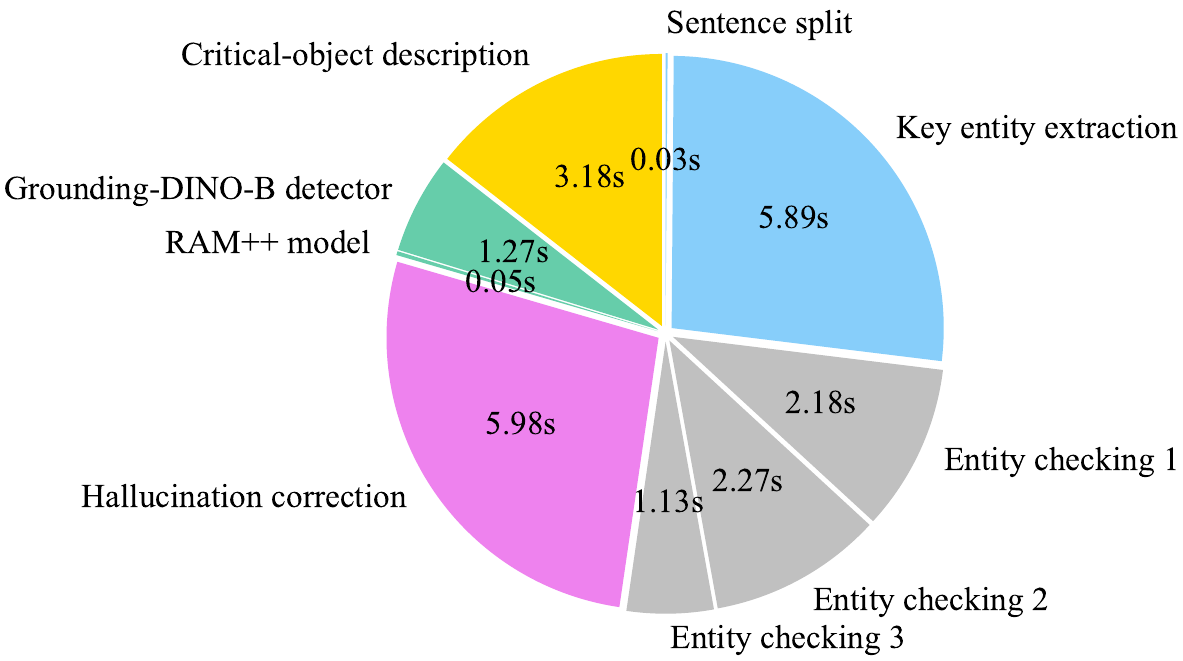}
	\caption{Inference time of each module in the HCOENet, with a total time of about 20 seconds. ``Entity checking 1" refers the BLIP-2-$\text{FlanT5}_{\mathrm{XXL}}$ method, ``Entity checking 2" refers the InstructBLIP-$\text{FlanT5}_{\mathrm{XXL}}$ method, and ``Entity checking 3" refers the InstructBLIP-Vicuna-13B method.}
	\label{Fig_inference_time}
\end{figure}

The inference time of each module in the HCOENet is detailed in Fig.~\ref{Fig_inference_time}, and the total processing time for each image is approximately 21.98 seconds. Among all stages, the most time-intensive are key entity extraction and hallucination correction, collectively accounting for 54\% of the total time. In these stages, the large language model Llama-3.1-8B is called for textual integration, where long prompts with embedded examples are contained. As a result, substantial time is required to comprehend the input prompts and generate outputs in the specified format. Besides, the entity cross-checking stage also consumes a considerable amount of time. Specifically, BLIP-2-$\text{FlanT5}_{\mathrm{XXL}}$ and InstructBLIP-$\text{FlanT5}_{\mathrm{XXL}}$ require 2.18 seconds and 2.27 seconds, respectively, as both models evaluate each extracted entity word. In contrast, InstructBLIP-Vicuna-13B takes only 1.13 seconds because it evaluates only entities with conflicting results from the first two models. This cross-checking approach significantly reduces the calculations and shortens the processing time.

The fastest stages in the HCOENet are sentence splitting and image tagging using the RAM++ method, which account for just 0.36\% of the total time. Unlike the Grounding-DINO-B model, which needs to recognize and locate targets, the RAM++ model focuses solely on object recognition, resulting in the Grounding-DINO-B model taking 25 times longer than RAM++. Especially, the time required for the RAM++ model, the Grounding-DINO-B detector, and the object description step relies on the complexity of the given image. For images with fewer critical targets, the required processing time is significantly shorter. Finally, the description merging stage is very fast, taking less than 0.001 seconds, and is negligible in the figure above.

\subsubsection{Self-Collected Images in Campus}   

\begin{figure*}[!t]
	\centering
	\includegraphics[width=0.95\linewidth]{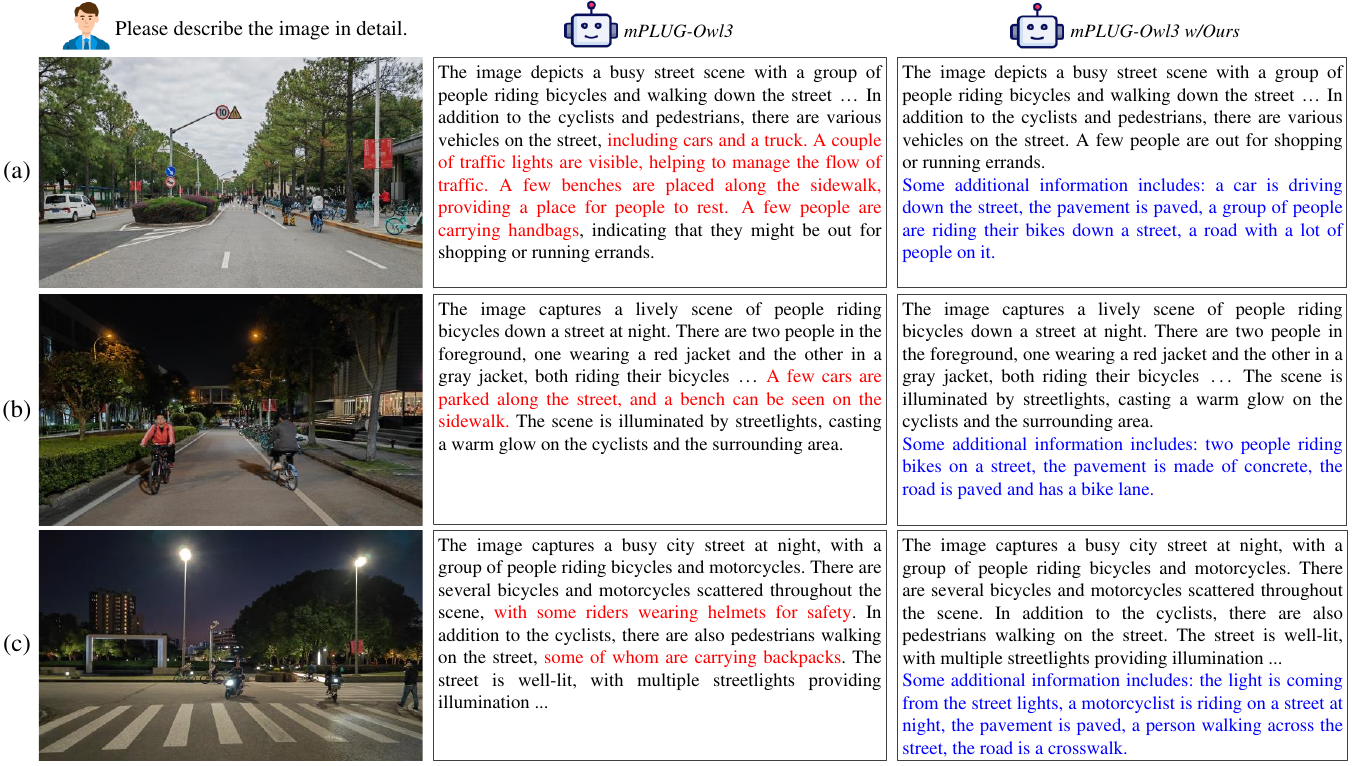}
	\caption{Hallucination correction cases of campus scene images on the mPLUG-Owl3 model. Hallucinatory texts are highlighted in red, and texts generated by critical-object enhancement framework are highlighted in blue.}
	\label{Fig_campus_results}
\end{figure*}

We collect some campus scene images for model inference, as shown in Fig.~\ref{Fig_campus_results}. For these images, the multimodal method mPLUG-Owl3 is employed to generate detailed descriptions using the prompt of ``\textit{Please describe the image in detail}". Fig.~\ref{Fig_campus_results}(a) is taken at noon, featuring many students walking or cycling on the road, along with many parked shared bicycles, all of which can be clearly described by the LVLM. However, the generated texts still contain several errors, such as hallucinations of trucks and benches, misidentifying traffic signs as traffic lights, and incorrectly describing people as carrying handbags. When using the proposed correction method, BLIP-2-$\text{FlanT5}_{\mathrm{XXL}}$ and InstructBLIP-$\text{FlanT5}_{\mathrm{XXL}}$ identify that trucks, traffic lights, benches, and handbags are not present in the scene. Subsequently, the Llama model eliminates these erroneous elements from the descriptions. This process effectively removes hallucinatory information, resulting in a more accurate textual representation of the image.

Fig.~\ref{Fig_campus_results}(b) is taken in a dimly lit campus environment at night, with few traffic participants on the road, mainly several students riding bicycles. However, the dark lighting conditions present great challenges for large vision-language models, leading to errors such as misidentifying the wide stairs on the right as a bench. Additionally, the model incorrectly assumes that cars are present in the current scene. The hallucination likely stems from the statistic bias, as vehicles are among the most frequently appearing objects in traffic scenes. In the HCOENet, three entity cross-checking methods accurately detect and filter out the erroneous entity words \textit{car} and \textit{bench}, thereby refining the passage. Fig.~\ref{Fig_campus_results}(c) features a crosswalk illuminated by streetlights, which brighten the surrounding campus environment. Here, the response of the mPLUG-Owl3 model contains errors in some details, such as misidentifying motorcyclists as wearing helmets and wrongly stating that pedestrians in the image are carrying backpacks. These erroneous contents are all corrected in the refined description.

\section{Dataset Annotation}


In addition to eliminating semantic hallucinations, the proposed HCOENet can function as an image description annotator, as illustrated in Fig.~\ref{Fig_semantic_labels_CODA_nuscenes}. Specifically, when the initial response generated by the LVLM is corrected through HCOENet, the resulting description can serve as a semantic annotation for the corresponding image. Compared with the image annotation method of GPT-4o, HCOENet offers comparable accuracy while being cost-free.
Since each stage in the HCOENet is fully automated and does not require manual validation, this annotation method can be easily scaled to larger image datasets. Using this approach, we have constructed two novel semantic understanding datasets for the traffic scene, including CODA\_desc dataset with 9695 image-text pairs and nuScenes\_desc dataset with 40157 image-text pairs.
Beyond the two datasets, the proposed HCOENet is adaptable to a broader range of scenarios, facilitating the creation of additional semantic understanding datasets.

\section{Conclusions and Future Work}
In this paper, we tackle the challenge of object hallucinations existed in vision-language models and further leverage the proposed HCOENet to construct two traffic scene semantic understanding datasets CODA\_desc and nuScenes\_desc. 
As a training-free chain-of-thought correction method, HCOENet integrates multiple off-the-shelf models, enabling seamless adaptation to various LVLMs, including mPLUG-Owl, LLaVA, and InternVL model.
Besides, we conduct extensive experiments on the POPE benchmark under three negative sampling settings, achieving significant improvements particularly in recall and F1-score metrics. Compared with the state-of-the-art GPT-4o model, our method obtains high-quality image descriptions with comparable performance at a significantly lower cost. Furthermore, some self-collected campus images are employed to evaluate the model's generalization capability, further demonstrating the practical application value of our approach in eliminating object hallucinations.

In the future, we aim to extend the HCOENet in the following directions: 1) Exploring more multimodal models in the entity cross-checking stage. 2) Expanding the model to address attribute and relation hallucinations.


\bibliographystyle{IEEEtran}
\bibliography{refer_hallucination}

\begin{thebibliography}{10}
\providecommand{\url}[1]{#1}
\csname url@samestyle\endcsname
\providecommand{\newblock}{\relax}
\providecommand{\bibinfo}[2]{#2}
\providecommand{\BIBentrySTDinterwordspacing}{\spaceskip=0pt\relax}
\providecommand{\BIBentryALTinterwordstretchfactor}{4}
\providecommand{\BIBentryALTinterwordspacing}{\spaceskip=\fontdimen2\font plus
\BIBentryALTinterwordstretchfactor\fontdimen3\font minus
  \fontdimen4\font\relax}
\providecommand{\BIBforeignlanguage}[2]{{%
\expandafter\ifx\csname l@#1\endcsname\relax
\typeout{** WARNING: IEEEtran.bst: No hyphenation pattern has been}%
\typeout{** loaded for the language `#1'. Using the pattern for}%
\typeout{** the default language instead.}%
\else
\language=\csname l@#1\endcsname
\fi
#2}}
\providecommand{\BIBdecl}{\relax}
\BIBdecl

\bibitem{TITS_1}
C.~Yang, K.~Zhuang, M.~Chen, H.~Ma, X.~Han, T.~Han, C.~Guo, H.~Han, B.~Zhao,
  and Q.~Wang, ``Traffic sign interpretation via natural language
  description,'' \emph{IEEE Trans. Intell. Transp. Syst.}, 2024.

\bibitem{TITS_2}
M.~Z. Hasan, J.~Chen, J.~Wang, M.~S. Rahman, A.~Joshi, S.~Velipasalar,
  C.~Hegde, A.~Sharma, and S.~Sarkar, ``Vision-language models can identify
  distracted driver behavior from naturalistic videos,'' \emph{IEEE Trans.
  Intell. Transp. Syst.}, 2024.

\bibitem{GPT_4}
J.~Achiam, S.~Adler, S.~Agarwal, L.~Ahmad, I.~Akkaya, F.~L. Aleman, D.~Almeida,
  J.~Altenschmidt, S.~Altman, S.~Anadkat \emph{et~al.}, ``Gpt-4 technical
  report,'' \emph{arXiv preprint arXiv:2303.08774}, 2023.

\bibitem{llava_o1}
G.~Xu, P.~Jin, L.~Hao, Y.~Song, L.~Sun, and L.~Yuan, ``Llava-o1: Let vision
  language models reason step-by-step,'' \emph{arXiv preprint
  arXiv:2411.10440}, 2024.

\bibitem{TITS_3}
C.~Liu, X.~Zhang, F.~Chang, S.~Li, P.~Hao, Y.~Lu, and Y.~Wang, ``Traffic
  scenario understanding and video captioning via guidance attention captioning
  network,'' \emph{IEEE Trans. Intell. Transp. Syst.}, 2023.

\bibitem{TITS_4}
Y.~Feng, W.~Hua, and Y.~Sun, ``Nle-dm: Natural-language explanations for
  decision making of autonomous driving based on semantic scene
  understanding,'' \emph{IEEE Trans. Intell. Transp. Syst.}, vol.~24, no.~9,
  pp. 9780--9791, 2023.

\bibitem{TITS_5}
M.~Movahedi and J.~Choi, ``The crossroads of llm and traffic control: A study
  on large language models in adaptive traffic signal control,'' \emph{IEEE
  Trans. Intell. Transp. Syst.}

\bibitem{Adapt}
B.~Jin, X.~Liu, Y.~Zheng, P.~Li, H.~Zhao, T.~Zhang, Y.~Zheng, G.~Zhou, and
  J.~Liu, ``Adapt: Action-aware driving caption transformer,'' in \emph{2023
  IEEE International Conference on Robotics and Automation (ICRA)}, 2023, pp.
  7554--7561.

\bibitem{hallucination_survey}
Z.~Bai, P.~Wang, T.~Xiao, T.~He, Z.~Han, Z.~Zhang, and M.~Z. Shou,
  ``Hallucination of multimodal large language models: A survey,'' \emph{arXiv
  preprint arXiv:2404.18930}, 2024.

\bibitem{LRV_Instruction}
F.~Liu, K.~Lin, L.~Li, J.~Wang, Y.~Yacoob, and L.~Wang, ``Mitigating
  hallucination in large multi-modal models via robust instruction tuning,'' in
  \emph{Proc. Int. Conf. Learn. Represent. (ICLR)}, 2023.

\bibitem{hallucidoctor}
Q.~Yu, J.~Li, L.~Wei, L.~Pang, W.~Ye, B.~Qin, S.~Tang, Q.~Tian, and Y.~Zhuang,
  ``Hallucidoctor: Mitigating hallucinatory toxicity in visual instruction
  data,'' in \emph{Proc. IEEE Conf. Comput. Vis. Pattern Recognit. (CVPR)},
  2024, pp. 12\,944--12\,953.

\bibitem{HACL}
C.~Jiang, H.~Xu, M.~Dong, J.~Chen, W.~Ye, M.~Yan, Q.~Ye, J.~Zhang, F.~Huang,
  and S.~Zhang, ``Hallucination augmented contrastive learning for multimodal
  large language model,'' in \emph{Proc. IEEE Conf. Comput. Vis. Pattern
  Recognit. (CVPR)}, 2024, pp. 27\,036--27\,046.

\bibitem{RLHF_1}
Z.~Sun, S.~Shen, S.~Cao, H.~Liu, C.~Li, Y.~Shen, C.~Gan, L.-Y. Gui, Y.-X. Wang,
  Y.~Yang \emph{et~al.}, ``Aligning large multimodal models with factually
  augmented rlhf,'' \emph{arXiv preprint arXiv:2309.14525}, 2023.

\bibitem{BLIP2}
J.~Li, D.~Li, S.~Savarese, and S.~Hoi, ``Blip-2: Bootstrapping language-image
  pre-training with frozen image encoders and large language models,'' in
  \emph{Proc. Int. Conf. Mach. Learn. (ICML)}, 2023, pp. 19\,730--19\,742.

\bibitem{LLaVA_1.5}
H.~Liu, C.~Li, Y.~Li, and Y.~J. Lee, ``Improved baselines with visual
  instruction tuning,'' in \emph{Proc. IEEE Conf. Comput. Vis. Pattern
  Recognit. (CVPR)}, 2024, pp. 26\,296--26\,306.

\bibitem{mplug_owl2}
Q.~Ye, H.~Xu, J.~Ye, M.~Yan, A.~Hu, H.~Liu, Q.~Qian, J.~Zhang, and F.~Huang,
  ``mplug-owl2: Revolutionizing multi-modal large language model with modality
  collaboration,'' in \emph{Proc. IEEE Conf. Comput. Vis. Pattern Recognit.
  (CVPR)}, 2024, pp. 13\,040--13\,051.

\bibitem{VIGC}
B.~Wang, F.~Wu, X.~Han, J.~Peng, H.~Zhong, P.~Zhang, X.~Dong, W.~Li, W.~Li,
  J.~Wang \emph{et~al.}, ``Vigc: Visual instruction generation and
  correction,'' in \emph{Proc. AAAI Conf. Artif. Intell. (AAAI)}, vol.~38,
  no.~6, 2024, pp. 5309--5317.

\bibitem{huang2023survey}
L.~Huang, W.~Yu, W.~Ma, W.~Zhong, Z.~Feng, H.~Wang, Q.~Chen, W.~Peng, X.~Feng,
  B.~Qin \emph{et~al.}, ``A survey on hallucination in large language models:
  Principles, taxonomy, challenges, and open questions,'' \emph{ACM T. Inform.
  Syst.}, 2023.

\bibitem{Haloquest_eccv2024}
Z.~Wang, G.~Bingham, A.~W. Yu, Q.~V. Le, T.~Luong, and G.~Ghiasi, ``Haloquest:
  A visual hallucination dataset for advancing multimodal reasoning,'' in
  \emph{Proc. Eur. Conf. Comput. Vis. (ECCV)}, 2024, pp. 288--304.

\bibitem{IVE}
X.~He, L.~Wei, L.~Xie, and Q.~Tian, ``Incorporating visual experts to resolve
  the information loss in multimodal large language models,'' \emph{arXiv
  preprint arXiv:2401.03105}, 2024.

\bibitem{ESREAL}
M.~Kim, M.~Kim, J.~Bae, S.~Choi, S.~Kim, and B.~Chang, ``Exploiting semantic
  reconstruction to mitigate hallucinations in vision-language models,'' in
  \emph{Proc. Eur. Conf. Comput. Vis. (ECCV)}, 2025, pp. 236--252.

\bibitem{VDGD}
S.~Ghosh, C.~K.~R. Evuru, S.~Kumar, U.~Tyagi, O.~Nieto, Z.~Jin, and D.~Manocha,
  ``Vdgd: Mitigating lvlm hallucinations in cognitive prompts by bridging the
  visual perception gap,'' \emph{arXiv preprint arXiv:2405.15683}, 2024.

\bibitem{M3ID}
A.~Favero, L.~Zancato, M.~Trager, S.~Choudhary, P.~Perera, A.~Achille,
  A.~Swaminathan, and S.~Soatto, ``Multi-modal hallucination control by visual
  information grounding,'' in \emph{Proc. IEEE Conf. Comput. Vis. Pattern
  Recognit. (CVPR)}, 2024, pp. 14\,303--14\,312.

\bibitem{VCoder}
J.~Jain, J.~Yang, and H.~Shi, ``Vcoder: Versatile vision encoders for
  multimodal large language models,'' in \emph{Proc. IEEE Conf. Comput. Vis.
  Pattern Recognit. (CVPR)}, 2024, pp. 27\,992--28\,002.

\bibitem{woodpecker}
S.~Yin, C.~Fu, S.~Zhao, T.~Xu, H.~Wang, D.~Sui, Y.~Shen, K.~Li, X.~Sun, and
  E.~Chen, ``Woodpecker: Hallucination correction for multimodal large language
  models,'' \emph{arXiv preprint arXiv:2310.16045}, 2023.

\bibitem{LLaVA_1.6}
\BIBentryALTinterwordspacing
H.~Liu, C.~Li, Y.~Li, B.~Li, Y.~Zhang, S.~Shen, and Y.~J. Lee, ``Llava-next:
  Improved reasoning, ocr, and world knowledge,'' January 2024. [Online].
  Available: \url{https://llava-vl.github.io/blog/2024-01-30-llava-next/}
\BIBentrySTDinterwordspacing

\bibitem{InternVL_1.5}
Z.~Chen, W.~Wang, H.~Tian, S.~Ye, Z.~Gao, E.~Cui, W.~Tong, K.~Hu, J.~Luo, Z.~Ma
  \emph{et~al.}, ``How far are we to gpt-4v? closing the gap to commercial
  multimodal models with open-source suites,'' \emph{arXiv preprint
  arXiv:2404.16821}, 2024.

\bibitem{MiniGPT4}
D.~Zhu, J.~Chen, X.~Shen, X.~Li, and M.~Elhoseiny, ``Minigpt-4: Enhancing
  vision-language understanding with advanced large language models,''
  \emph{arXiv preprint arXiv:2304.10592}, 2023.

\bibitem{mplug_owl3}
J.~Ye, H.~Xu, H.~Liu, A.~Hu, M.~Yan, Q.~Qian, J.~Zhang, F.~Huang, and J.~Zhou,
  ``mplug-owl3: Towards long image-sequence understanding in multi-modal large
  language models,'' \emph{arXiv preprint arXiv:2408.04840}, 2024.

\bibitem{InstructBLIP}
W.~Dai, J.~Li, D.~Li, A.~M.~H. Tiong, J.~Zhao, W.~Wang, B.~Li, P.~Fung, and
  S.~C.~H. Hoi, ``Instructblip: Towards general-purpose vision-language models
  with instruction tuning,'' in \emph{Proc. Adv. Neural Inf. Process. Syst.
  (NeurIPS)}, 2023.

\bibitem{Qwen_vl}
J.~Bai, S.~Bai, S.~Yang, S.~Wang, S.~Tan, P.~Wang, J.~Lin, C.~Zhou, and
  J.~Zhou, ``Qwen-vl: A versatile vision-language model for understanding,
  localization, text reading, and beyond,'' 2023.

\bibitem{Recaption}
L.~Wang, J.~He, S.~Li, N.~Liu, and E.-P. Lim, ``Mitigating fine-grained
  hallucination by fine-tuning large vision-language models with caption
  rewrites,'' in \emph{Proc. Int. Conf. Multi. Model. (ICMM)}, 2024, pp.
  32--45.

\bibitem{LION}
G.~Chen, L.~Shen, R.~Shao, X.~Deng, and L.~Nie, ``Lion: Empowering multimodal
  large language model with dual-level visual knowledge,'' in \emph{Proc. IEEE
  Conf. Comput. Vis. Pattern Recognit. (CVPR)}, 2024, pp. 26\,540--26\,550.

\bibitem{IBD}
L.~Zhu, D.~Ji, T.~Chen, P.~Xu, J.~Ye, and J.~Liu, ``Ibd: Alleviating
  hallucinations in large vision-language models via image-biased decoding,''
  \emph{arXiv preprint arXiv:2402.18476}, 2024.

\bibitem{HALC}
Z.~Chen, Z.~Zhao, H.~Luo, H.~Yao, B.~Li, and J.~Zhou, ``Halc: Object
  hallucination reduction via adaptive focal-contrast decoding,'' \emph{arXiv
  preprint arXiv:2403.00425}, 2024.

\bibitem{LURE}
Y.~Zhou, C.~Cui, J.~Yoon, L.~Zhang, Z.~Deng, C.~Finn, M.~Bansal, and H.~Yao,
  ``Analyzing and mitigating object hallucination in large vision-language
  models,'' \emph{arXiv preprint arXiv:2310.00754}, 2023.

\bibitem{VCD}
S.~Leng, H.~Zhang, G.~Chen, X.~Li, S.~Lu, C.~Miao, and L.~Bing, ``Mitigating
  object hallucinations in large vision-language models through visual
  contrastive decoding,'' in \emph{Proc. IEEE Conf. Comput. Vis. Pattern
  Recognit. (CVPR)}, 2024, pp. 13\,872--13\,882.

\bibitem{LogicCheckGPT}
J.~Wu, Q.~Liu, D.~Wang, J.~Zhang, S.~Wu, L.~Wang, and T.~Tan, ``Logical closed
  loop: Uncovering object hallucinations in large vision-language models,''
  \emph{arXiv preprint arXiv:2402.11622}, 2024.

\bibitem{Volcano}
S.~Lee, S.~H. Park, Y.~Jo, and M.~Seo, ``Volcano: mitigating multimodal
  hallucination through self-feedback guided revision,'' \emph{arXiv preprint
  arXiv:2311.07362}, 2023.

\bibitem{llama3}
A.~Dubey, A.~Jauhri, A.~Pandey, A.~Kadian, A.~Al-Dahle, A.~Letman, A.~Mathur,
  A.~Schelten, A.~Yang, A.~Fan \emph{et~al.}, ``The llama 3 herd of models,''
  \emph{arXiv preprint arXiv:2407.21783}, 2024.

\bibitem{RAM++}
X.~Huang, Y.-J. Huang, Y.~Zhang, W.~Tian, R.~Feng, Y.~Zhang, Y.~Xie, Y.~Li, and
  L.~Zhang, ``Open-set image tagging with multi-grained text supervision,''
  \emph{arXiv e-prints}, pp. arXiv--2310, 2023.

\bibitem{GroundingDINO}
S.~Liu, Z.~Zeng, T.~Ren, F.~Li, H.~Zhang, J.~Yang, C.~Li, J.~Yang, H.~Su,
  J.~Zhu \emph{et~al.}, ``Grounding dino: Marrying dino with grounded
  pre-training for open-set object detection,'' \emph{arXiv preprint
  arXiv:2303.05499}, 2023.

\bibitem{CODA}
K.~Li, K.~Chen, H.~Wang, L.~Hong, C.~Ye, J.~Han, Y.~Chen, W.~Zhang, C.~Xu,
  D.-Y. Yeung \emph{et~al.}, ``Coda: A real-world road corner case dataset for
  object detection in autonomous driving,'' in \emph{Proc. Eur. Conf. Comput.
  Vis. (ECCV)}, 2022, pp. 406--423.

\bibitem{mplug_owl}
Q.~Ye, H.~Xu, G.~Xu, J.~Ye, M.~Yan, Y.~Zhou, J.~Wang, A.~Hu, P.~Shi, Y.~Shi
  \emph{et~al.}, ``mplug-owl: Modularization empowers large language models
  with multimodality,'' \emph{arXiv preprint arXiv:2304.14178}, 2023.

\bibitem{POPE}
Y.~Li, Y.~Du, K.~Zhou, J.~Wang, W.~X. Zhao, and J.-R. Wen, ``Evaluating object
  hallucination in large vision-language models,'' \emph{arXiv preprint
  arXiv:2305.10355}, 2023.

\bibitem{SEEM}
X.~Zou, J.~Yang, H.~Zhang, F.~Li, L.~Li, J.~Wang, L.~Wang, J.~Gao, and Y.~J.
  Lee, ``Segment everything everywhere all at once,'' vol.~36, 2024.

\bibitem{YOLOv9}
C.-Y. Wang, I.-H. Yeh, and H.-Y.~M. Liao, ``Yolov9: Learning what you want to
  learn using programmable gradient information,'' \emph{arXiv preprint
  arXiv:2402.13616}, 2024.

\bibitem{YOLOv10}
A.~Wang, H.~Chen, L.~Liu, K.~Chen, Z.~Lin, J.~Han, and G.~Ding, ``Yolov10:
  Real-time end-to-end object detection,'' \emph{arXiv preprint
  arXiv:2405.14458}, 2024.

\bibitem{YOLOv11}
R.~Khanam and M.~Hussain, ``Yolov11: An overview of the key architectural
  enhancements,'' \emph{arXiv preprint arXiv:2410.17725}, 2024.

\bibitem{Tag2text}
X.~Huang, Y.~Zhang, J.~Ma, W.~Tian, R.~Feng, Y.~Zhang, Y.~Li, Y.~Guo, and
  L.~Zhang, ``Tag2text: Guiding vision-language model via image tagging,''
  \emph{arXiv preprint arXiv:2303.05657}, 2023.

\bibitem{RAM}
Y.~Zhang, X.~Huang, J.~Ma, Z.~Li, Z.~Luo, Y.~Xie, Y.~Qin, T.~Luo, Y.~Li, S.~Liu
  \emph{et~al.}, ``Recognize anything: A strong image tagging model,''
  \emph{arXiv preprint arXiv:2306.03514}, 2023.

\end{thebibliography}

\end{CJK*}
\end{document}